\documentclass{article}

 \usepackage[main, final]{neurips_2026}

\usepackage[utf8]{inputenc} 
\usepackage[T1]{fontenc}    
\usepackage{hyperref}       
\usepackage{url}            
\usepackage{booktabs}       
\usepackage{amsfonts}       
\usepackage{nicefrac}       
\usepackage{microtype}      
\usepackage{xcolor}         
\usepackage{graphicx}
\usepackage{amsmath}
\usepackage{subcaption}
\usepackage[most]{tcolorbox}

\title{Revealing the Gap in Human and VLM Scene Description through Counterfactual Semantic Saliency}

\author{
  \textbf{Ziqi Wen}\textsuperscript{1} \quad
  \textbf{Parsa Madinei}\textsuperscript{1,2}\quad
  \textbf{Miguel P. Eckstein}\textsuperscript{1,2} \\
  \textsuperscript{1}Department of Computer Science, University of California, Santa Barbara\\
  \textsuperscript{2}Department of Psychological and Brain Sciences, University of California, Santa Barbara\\
  \small\texttt{\{ziqiwen, madinei, migueleckstein\}@ucsb.edu }
}

\begin{document}

\maketitle

\begin{abstract}
Evaluating whether large vision-language models (VLMs) align with human perception for high-level semantic scene comprehension remains a challenge. Traditional white-box interpretability methods are inapplicable to closed-source architectures and passive metrics fail to isolate causal features. We introduce Counterfactual Semantic Saliency (CSS). This black-box, model-agnostic framework quantifies the importance of objects by measuring the semantic shift induced by their causal ablation from a scene. To evaluate AI-human semantic alignment, we tested prominent VLMs against a human psychophysics baseline comprising 16,603 valid responses across 307 complex natural scenes and 1,306 high-fidelity counterfactual variants. Our analysis reveals a pervasive scene comprehension gap: models exhibit an over-reliance (relative to humans) on object size, object centrality, and low-level saliency of objects.  In contrast, models are less sensitive to the presence of people in the scenes than our human participants when describing the images.  The over-reliance on object size is a primary driver explaining variations in model-human semantic divergence. Code and data are available at \href{https://github.com/starsky77/Counterfactual-Semantic-Saliency-Data}{https://github.com/starsky77/Counterfactual-Semantic-Saliency-Data}.

\end{abstract}

\begin{figure}[ht]
    \centering
    \includegraphics[width=\linewidth]{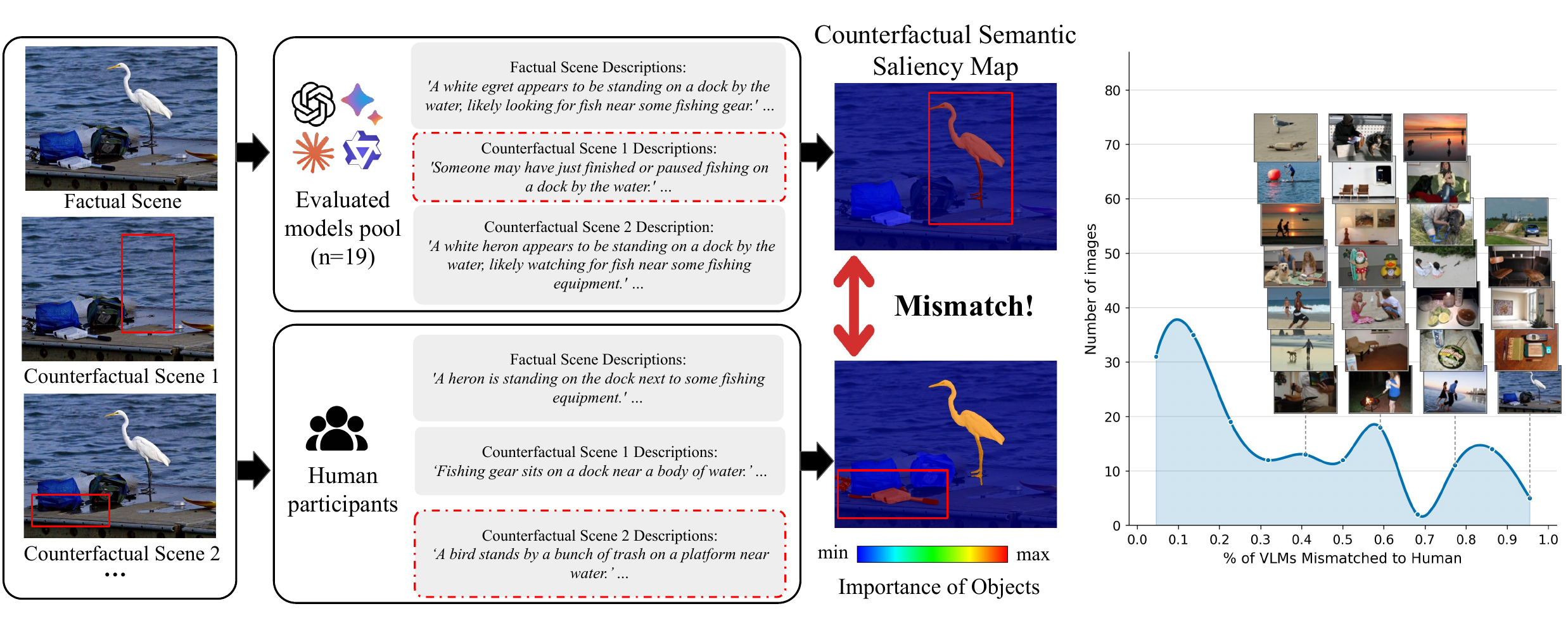}
    \caption{\textbf{Counterfactual Semantic Saliency (CSS) reveals a pervasive scene comprehension gap between humans and vision-language models (VLMs).} CSS quantifies the importance of an object by ablating it from a factual scene and measuring the semantic shift in the resulting descriptions. The object causing the maximal semantic shift is considered the most critical to scene perception. The example illustrates a mismatch between ALL 19 models and humans. The rightmost plot illustrates the percentage of the evaluated VLMs that failed to identify the human-consensus critical object for a given scene.}

    \label{fig:teaser}
\end{figure}

\section{Introduction}

As deep neural networks (DNNs) achieve superior performance, verifying whether these models exhibit perceptual strategies aligned with humans has become a critical challenge. Studies in AI-human alignment have revealed a persistent divergence in these strategies: even when achieving comparable accuracy on benchmark tasks, DNNs and human observers often rely on different features. Compared with humans, models tend to show over-reliance on local textures \citep{geirhos2018imagenet,geirhos2020shortcut}, subtle pixel patterns \citep{ilyas2019adversarial}, other localized cues that are highly inconsistent with holistic human perception \citep{fel2022harmonizing, wen2023does}, or even ignoring the spatial relationship of visual elements\citep{yuksekgonul2022and}. While scaling parameters and training data have been shown to induce emergent alignment \citep{geirhos2021partial, gokce2024scaling, li2025local}, these evaluations have primarily focused on isolated, simplified tasks such as basic object classification. Consequently, understanding the gap between modern Vision-Language Models (VLMs) and human observers in the context of scene comprehension, which requires integrating multiple objects, spatial contexts, and complex semantic relationships, remains largely underexplored.

To investigate this alignment gap, the driving factors behind model decisions need to be understood. However, the scale, complexity, and frequently closed-source nature of modern VLMs have made many traditional "white-box" interpretability techniques impractical. Evaluating the mechanisms of these architectures requires a shift toward black-box, model-agnostic strategies. Causal intervention frameworks \citep{ullman2016atoms, goyal2019counterfactual, spanos2025v,ding2025explanation} have emerged as a powerful solution; by systematically manipulating input variables, such as ablating specific visual elements, we can directly measure the downstream effects and isolate the causal weight of individual scene components.

Building upon this paradigm, we propose Counterfactual Semantic Saliency (CSS), a novel, black-box, and model-agnostic methodology designed to evaluate object importance in scene perception. CSS enables us to systematically investigate \textit{whether modern VLMs perceive and prioritize elements in complex scenes in alignment with human observers.} To compare model and human semantic alignment, we constructed a human baseline using complex, multi-object images from a widely adopted human saliency dataset \citep{xu2014predicting}. Using state-of-the-art generative inpainting with Nano Banana 2 \citep{team2023gemini}, we created 307 factual scenes and 1,306 high-fidelity counterfactual variants by systematically ablating 3--6 visual elements per scene. We then conducted a large-scale human psychophysics study, collecting 16,603 valid textual descriptions across all 1,613 images from 227 independent participants.

Pervasive gaps between models and human scene comprehension were found through statistical analysis. Notably, every evaluated VLM failed to reach human-human consistency levels when predicting the most critical object for scene comprehension or the object importance ranking. Furthermore, we identified divergences in the underlying attribute sensitivity during scene viewing: whereas humans focus more on the presence of people, models heavily rely on bottom-up attributes, including object size, object centrality, and low-level saliency. Among these divergent factors, we demonstrate that exaggerated VLM sensitivity to object size is the dominant factor contributing to the human-AI semantic alignment gap.

The primary contributions of this work are as follows:
\begin{itemize}
    \item We introduce Counterfactual Semantic Saliency (CSS), a black-box and model-agnostic methodology to systematically quantify the importance of objects in scene description, overcoming the limitations of traditional white-box interpretability methods.
    \item We establish a human baseline by conducting a large-scale human psychophysics study, yielding a comprehensive benchmark dataset comprising 16,603 valid responses across 307 factual scenes and 1,306 counterfactual variants from 227 independent participants.
    \item We provide the first systematic quantification of the semantic alignment gap between modern VLMs and human observers in scene description, revealing a pervasive divergence in the sensitivity to 4 object attributes (size, centrality, low-level saliency, person-non-person) between VLMs and humans.
    \item Through robust statistical analysis, we discover the major driving factor of this scene comprehension gap, demonstrating that an excessive reliance on size in modern architectures contributes the most to their divergence from humans.
\end{itemize}

\section{Related Works}

Historically, understanding the decision-making processes of visual models like CNNs relied heavily on white-box methods, ranging from gradient-based \citep{selvaraju2017grad} and activation-based \citep{bau2017network} internal analyses to early black-box probing \citep{ribeiro2016should, lundberg2017unified}. As Vision Transformers become the dominant vision encoder architecture, visualizing attention maps provides cues about the features that contribute to a model's output \citep{chefer2021generic,leem2024attention,jo2025gmar}. However, the immense complexity of modern VLMs makes these straightforward techniques no longer sufficient to fully capture the true causal features driving a model's generation \citep{jain2019attention, chefer2021transformer}. While black-box analytical methods have long existed, previous approaches to complex visual tasks have faced significant obstacles. When altering image inputs for causal intervention, researchers either use an oversimplified method that blocks different scene regions \citep{murlidaran2026evolution}, or, when altering the scene, need to ensure that the modified image quality does not introduce confounding artifacts \citep{spanos2025v}, or they have to collect naturally occurring image pairs that were visually similar but semantically distinct \citep{tong2024eyes}. Although traditional inpainting models could rapidly remove objects from scenes \citep{li2022mat, rombach2022high, suvorov2022resolution, zhuang2024task}, the visible traces left behind by these modifications were non-negligible.

However, recent advancements in generative AI have made high-quality, flawless, and fine-grained image editing viable \citep{peebles2023scalable, team2023gemini, wallace2024diffusion, xie2024sana}. SOTA closed-source models are capable of generating images that are nearly indistinguishable to human observers, and the maturation of image-to-image generation has enabled highly controllable editing \citep{fu2023guiding, rout2024semantic,ju2024brushnet}. This technological leap serves as the foundational premise for our work. Counterfactual Semantic Saliency (CSS) leverages the most advanced generative models available (Nano Banana 2) to produce high-fidelity counterfactual images, allowing us to systematically verify whether a model's semantic shifts align with human perception. Crucially, the exceptional visual quality ensures that neither the evaluated models nor the human participants detected generative artifacts, providing the necessary foundation for an unbiased comparison.

\section{Method}

\subsection{Counterfactual Semantic Saliency (CSS)}

\begin{figure}[ht]
    \centering
    \includegraphics[width=\linewidth]{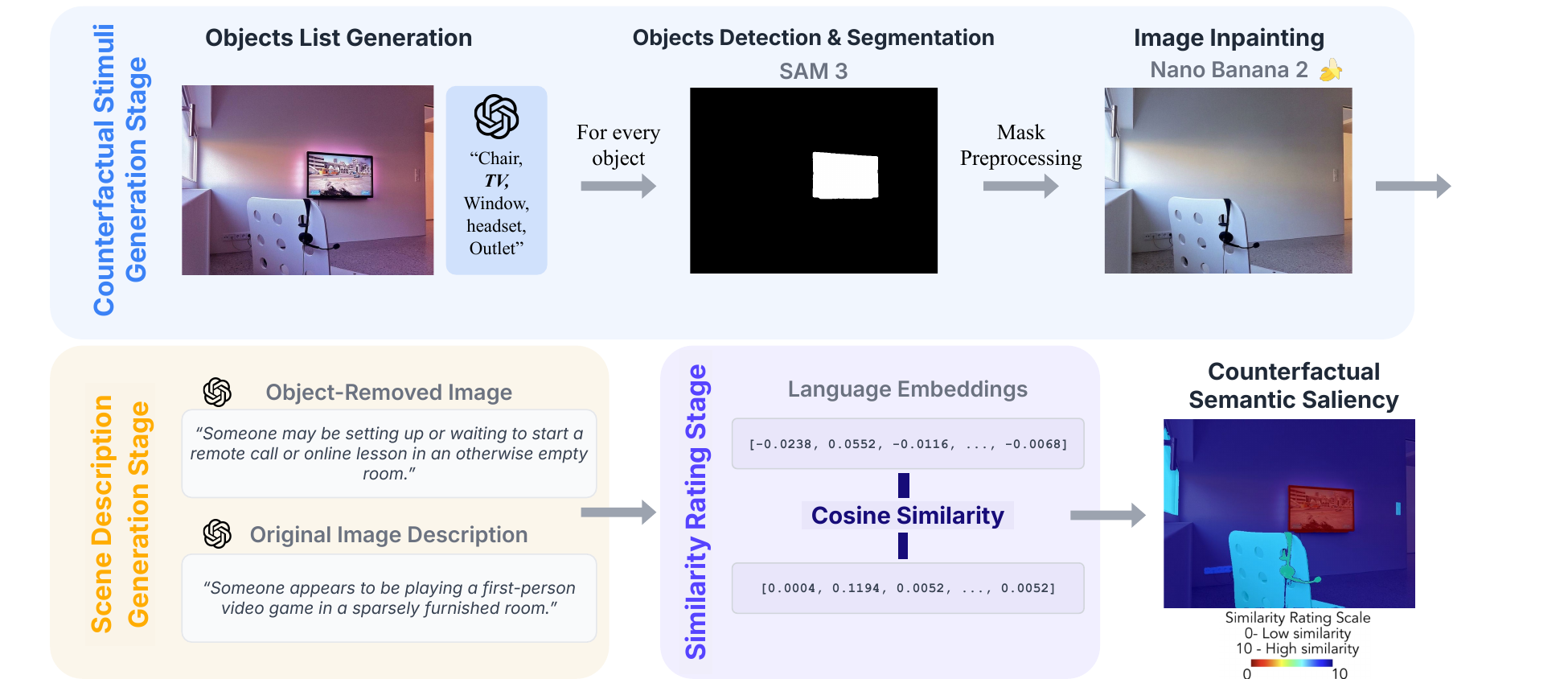}
    \caption{\textbf{Workflow of Counterfactual Semantic Saliency.} (1) Counterfactual Stimuli Generation Stage: A large vision-language model generates a list of object names in the scene. For each identified object, SAM3 generates a corresponding segmentation. Following mask preprocessing, an image inpainting model (Nano Banana 2) removes the target object. In the illustration, the TV, along with the pink LED light, is removed from the scene. (2) Scene Description Stage: Descriptions are sampled from the target evaluation model for both the original image and all counterfactual variants. (3) Similarity Rating Stage: Counterfactual Semantic Saliency is quantified as one minus the mean pairwise cosine similarity between the sentence embeddings of the factual scene and counterfactual scene descriptions.}
    \label{fig:method}
\end{figure}

\subsubsection{Counterfactual Stimuli Generation}

Counterfactual Semantic Saliency (CSS) defines object importance by measuring the semantic shift induced by its removal as a counterfactual intervention. To generate these counterfactual stimuli, the initial stage consists of the following three steps: object list generation, object segmentation, and image inpainting, as illustrated in Fig.\ref{fig:method}.

\textbf{Object list generation:} We applied GPT4o\citep{achiam2023gpt} to generate a list of object names given the scene. Objects that could be considered part of the background (e.g., sky, sea, wall, ground) are explicitly excluded; the removal of those objects would induce disproportionate semantic shifts or obvious hallucination due to their spatial dominance. We generated $N$ distinct responses via multinomial sampling and merged them using GPT-4o to eliminate duplicated labels. Detailed prompts are provided in the Appendix. 

\textbf{Object Segmentation \& Mask Preprocessing:} Segmentation masks were generated using the text-prompted segmentation model, SAM3 \citep{carion2025sam},  with a confidence threshold of 0.4. To isolate meaningful visual elements before object removal, we applied a four-step preprocessing pipeline. First, duplicated masks are removed by checking all possible pairs of masks whose Intersection over Union (IoU) is greater than 95\%. Second, guided by principles of perceptual grouping \citep{palmer1999vision}, we merged spatially proximate masks (distance < 30 pixels) that have identical labels into a singular mask. This ensures that higher-order structural elements (e.g., clustered candles forming a heart shape) are treated as unified conceptual entities rather than disjointed features, and individual masks were discarded after merging. Third, we discarded masks occupying >30\% of the total image area to prevent severe hallucination and context collapse associated with removing large structural components. Finally, a dilation was applied to ensure complete target coverage before the inpainting process. 

 \textbf{Image Inpainting:}  While conventional inpainting models such as MAT \citep{li2022mat} can remove the object effectively and efficiently, they often introduce artifacts when handling large masks or complex occlusions. Therefore, we utilized the SOTA generative model Nano Banana 2 \citep{team2023gemini} to generate high-fidelity, artifact-free counterfactual images. To control for model hallucinations or any undefined behavior introduced by the model, all generated stimuli underwent rigorous human validation. Fourteen independent annotators (recruited via Prolific) verified the successful ablation of the target objects. Images flagged for incomplete removal were manually reviewed and excluded from the final experimental dataset. Detailed validation protocols are provided in the Appendix.

\subsubsection{Measuring Semantic Shift with CSS scores}

We formalize Counterfactual Semantic Saliency (CSS) through a causal intervention framework. Let $\mathcal{I}$ denote the original scene, $o$ the target objects, and $\mathcal{M}$ the target vision-language model evaluated for scene description. We represent the targeted ablation of $o$ using \textit{do}-operator as the intervention $do(o = \emptyset)$, which yields the counterfactual stimulus $\mathcal{I}_{do(o=\emptyset)}$. To account for the variability in the model's generated descriptions, we sample a set of $N$ descriptions (denoted by $d$) for both the factual condition, $D = \{d_1, \dots, d_N\} \sim \mathcal{M}(\mathcal{I})$, and the counterfactual condition, $D' = \{d'_1, \dots, d'_N\} \sim \mathcal{M}(\mathcal{I}_{do(o=\emptyset)})$. Let $\mathcal{E}(\cdot)$ denote the sentence embedding function, which we implement using the open-source sentence embedding model Jasper \citep{zhang2024jasper}. The CSS score of object $o$ is defined as one minus the expected cosine similarity between the semantic representations of the factual and counterfactual descriptions. We approximate this semantic shift as  one minus the mean pairwise cosine similarity between the two sampled sets:
$$ \text{CSS}(o) = 1- \frac{1}{N^2} \sum_{j=1}^{N} \sum_{k=1}^{N} \frac{\mathcal{E}(d_j) \cdot \mathcal{E}(d'_k)}{\|\mathcal{E}(d_j)\| \|\mathcal{E}(d'_k)\|} $$
where $ N=5$ for all the models. Under this formulation, CSS score acts as a metric of perceptual importance.  For each scene, the semantically critical object is defined as the object that, when removed, impacts the scene description the most (highest CSS score). Finally, these object-level CSS scores are mapped back onto their corresponding segmentation masks to construct a top-down counterfactual semantic saliency map for the entire visual scene.

\subsection{Dataset}

We evaluated our framework using the Object and Semantic Images and Eye-tracking (OSIE) dataset \citep{xu2014predicting}, a standard benchmark in visual saliency research, consisting of 700 complex natural scenes with paired free-viewing eye-tracking data \citep{de2019individual}. The rich, multi-object composition of these foregrounds provides an ideal testbed for targeted counterfactual analysis, enabling the systematic isolation of semantic contributions. To maintain a controlled density of semantic elements and prevent overly cluttered or sparse interventions, we filtered the initial pool of images to include only scenes containing between 3 and 6 counterfactual variants (98 three-variant scenes, 87 four-variant scenes, 68 five-variant scenes, and 54 six-variant scenes). Following the generation of counterfactual stimuli and human validation of the inpainting quality, our final dataset contains 307 unique factual scenes and 1,306  counterfactual variants.

\subsection{Human Psychophysics}
\begin{figure}[ht]
    \centering
    \includegraphics[width=0.9\linewidth]{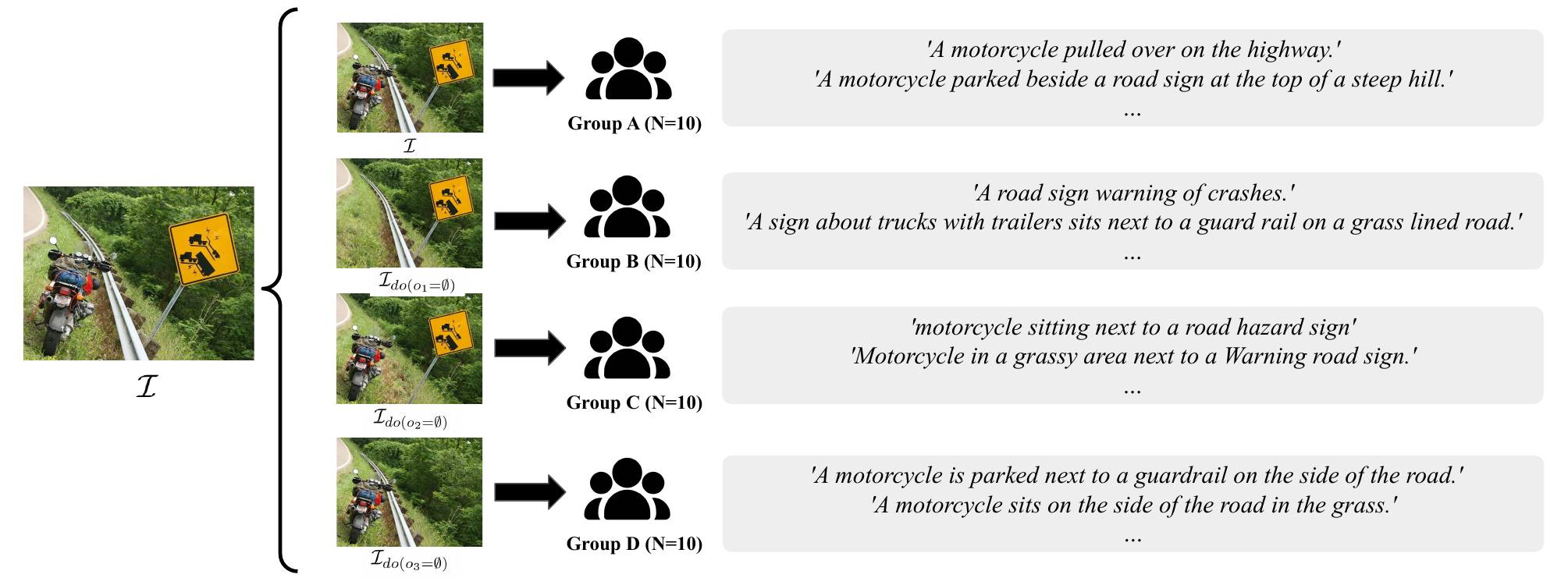}
    \caption{\textbf{Between-subjects experimental design for human psychophysics.} A scene containing $N$ target objects (1 factual image $\mathcal{I}$ and $N$ counterfactual variants $\mathcal{I}_{do(o_i=\emptyset)}$) was distributed across $N+1$ mutually exclusive stimulus sets. Each set was evaluated by an independent group of 10 participants, ensuring no individual was exposed to multiple variants of the same scene. We recruited 227 Subjects to complete a total of 16,665 trials.}
    \label{fig:human_study}
\end{figure}

To evaluate the alignment between vision-language models and human observers in scene comprehension, we established a human baseline for Counterfactual Semantic Saliency, with 227 subjects and 16,665 trials. Mirroring the pipeline used to evaluate the models, we collected human scene descriptions for all factual scenes and their corresponding counterfactual variants via the Prolific platform.

To prevent memory and carryover effects, it was strictly required that no single participant view multiple variants of the same scene. We therefore implemented a between-subjects experimental design, as illustrated in Fig.\ref{fig:human_study}. Factual scenes were first divided into four groups based on their target object count (ranging from 3 to 6 objects). For a scene containing \textit{N} target objects, the factual image and its \textit{N} counterfactual variants were distributed across \textit{N} + 1 mutually exclusive stimulus sets. Each stimulus set was then assigned to an independent group of 10 participants.\footnote{Incidental oversampling occurred for certain stimuli during crowdsourcing; all collected descriptions were kept in the final dataset.} This routing protocol ensured that we collected 10 independent textual descriptions for each unique stimulus, while strictly guaranteeing that no participant was exposed to more than one condition for any given scene. To filter low-quality responses that are oversimplified or irrelevant to the scene, we discard responses whose mean cosine similarity of sentence embeddings \citep{zhang2024jasper} with other human responses is below 0.4. Of the 16,665 trials, 62 (\textasciitilde0.3\%) are discarded in this process.

\section{Results}
\label{others}

\subsection{Gap between Human and Model Scene Comprehension}\label{sec:Perceptual_Gap}

\begin{figure}[htbp]
    \centering
    \begin{subfigure}[b]{0.48\textwidth}
        \centering
        \includegraphics[width=\textwidth]{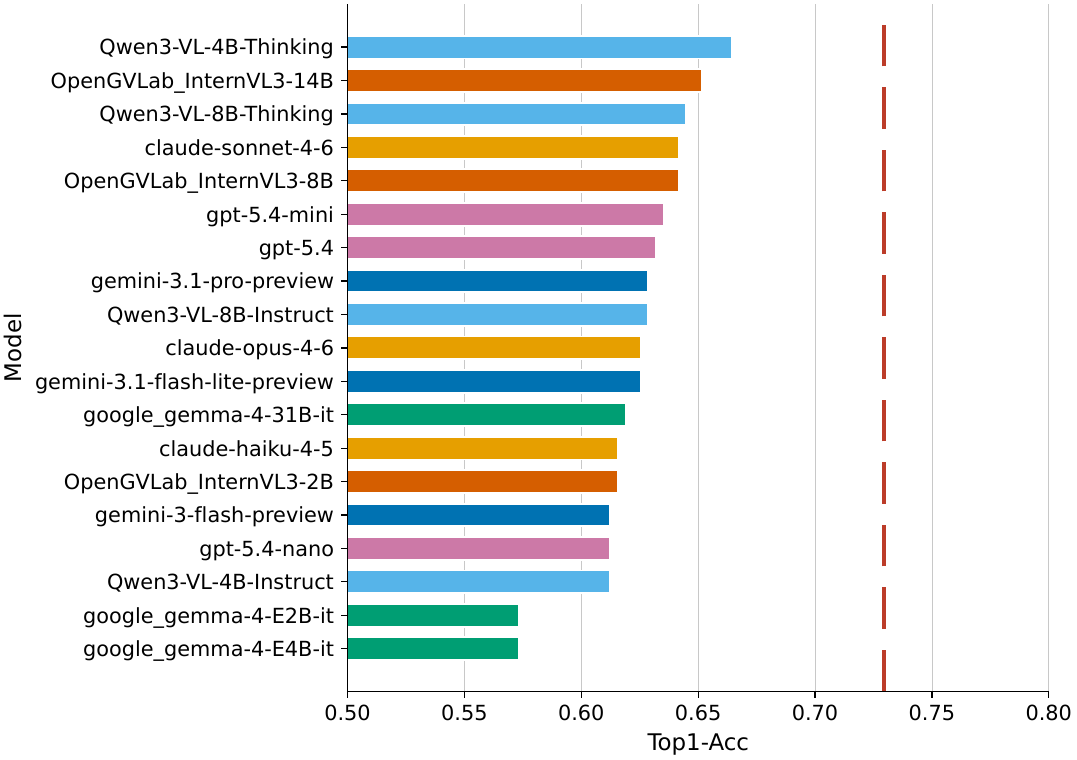}
        \caption{Top-1 Accuracy at predicting the critical object.}
        \label{fig:human_model_top1Acc}
    \end{subfigure}
    \hfill 
    \begin{subfigure}[b]{0.48\textwidth}
        \centering
        \includegraphics[width=\textwidth]{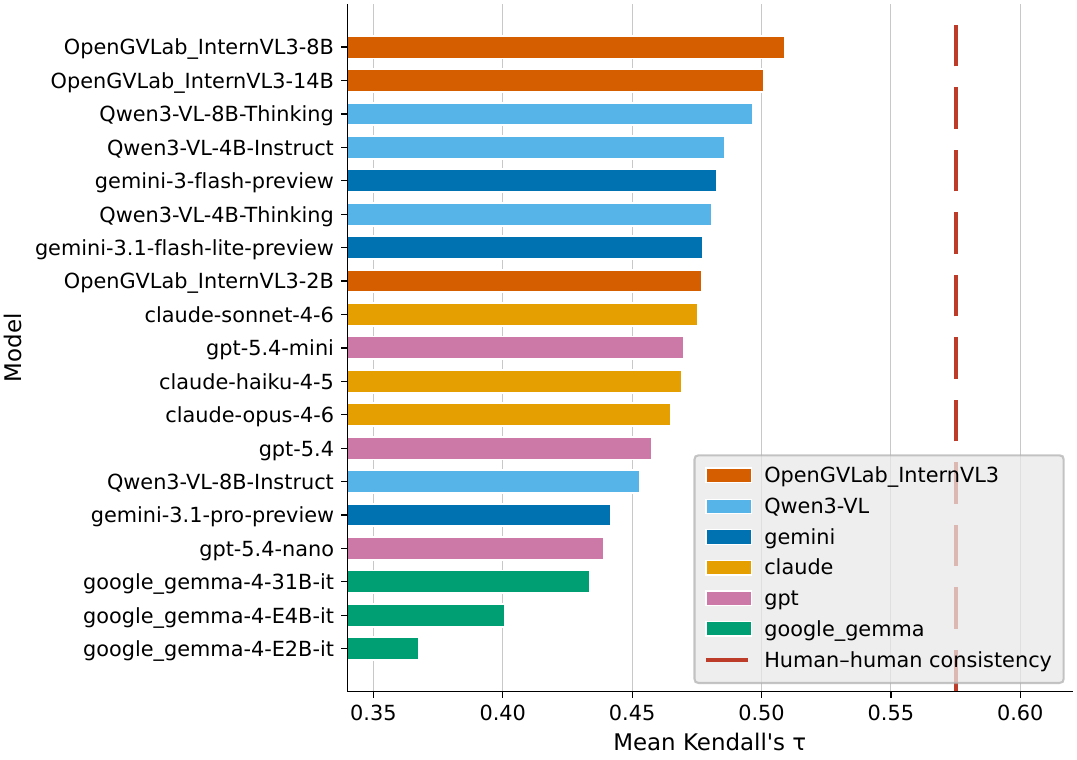}
        \caption{Mean Kendall's $\tau$ between models and humans.}
        \label{fig:human_model_tau}
    \end{subfigure}
    
    \caption{\textbf{Model-human alignment on scene description.} \textbf{(a)} Top-1 Accuracy representing the proportion of scenes where the evaluated model successfully identified the most semantically critical object, as determined by humans. \textbf{(b)} Mean Kendall's rank correlation coefficient ($\tau$) across all scenes, capturing the overall ordinal alignment of object importance hierarchies between models and humans. Full distributions of Kendall's $\tau$ for each model are provided in the Appendix. All the gaps here are significant with a one-tail bootstrap test ($p<.01$). }
    \label{fig:overall_eval}
\end{figure}

We quantified the alignment between human observers and all the commonly used open-source \citep{zhu2025internvl3,Kamath2025Gemma3T,bai2025qwen3} and closed-source 
\citep{singh2025openai,anthropic2024claude,comanici2025gemini}
vision-language models (n=19) using two primary metrics: Top-1 Accuracy and Kendall's rank correlation coefficient ($\tau$). Top-1 Accuracy measures the strict match rate, defined as the proportion of scenes where the model correctly identified the single most critical object (i.e., the target whose ablation induced the highest CSS score) as determined by the human baseline. Furthermore, we employed Kendall's $\tau$ to evaluate the structural alignment, measuring the ordinal correlation between the full rank-ordered hierarchies of object importance generated by the models versus humans.

To estimate human-human consistency, we partitioned the human data from 10 participants per counterfactual scene into two groups of five participant descriptions: one reference group and one comparison group. For the reference group, descriptions for counterfactual scenes from 5 participants were compared to 10 descriptions from factual scenes using cosine similarity. To mitigate individual subjective variance, we computed a human aggregate metric of CSS score, averaged across 50 description pairs' cosine similarity (5 counterfactual vs. 10 factual). All VLMs were subsequently evaluated against the reference group to ensure a controlled comparison. For human–human consistency baseline, we computed CSS maps from the comparison group and compared them with the CSS maps from the reference group using Top-1 accuracy and Kendall’s $\tau$ \footnote{Due to the incidental oversampling and filtering described previously, this comparison group occasionally contained between four and seven responses rather than five.}. As illustrated in Fig. \ref{fig:overall_eval}, a persistent and systematic gap exists between human and model performance in counterfactual scene comprehension. While the human-human consistency for Top-1 Accuracy reached 73\%, all evaluated models fell significantly short, ranging from 57\% to 65\%. This divergence is similarly reflected in the structural alignment metric: the human-human consistency achieved a mean Kendall's $\tau$ of 0.58, whereas model scores ranged from 0.37 to 0.51. Importantly, this divergence cannot be attributed to the choice of sentence embedding model, artifacts introduced by the inpainting model, or the variability of human agreement. We present further ablations in the Appendix confirming the robustness of this scene comprehension gap.

\subsection{Attribute Sensitivity in Models and Humans}

\subsubsection{Possible Hypotheses}

To identify the underlying factors driving the semantic shift resulting from object removal, we draw on established literature on human visual attention and DNN model bias. We investigated four primary attributes that are believed to influence CSS: size, centrality, low-level saliency, and whether the object is a person. We measured how these attributes influence object importance by correlating the CSS of removed objects with the objects' attributes across all 1,306 counterfactual scenes, referred to as "sensitivity to X" in the following. Spearman correlation is applied for numerical attributes (e.g., object size), and point-biserial correlation is applied for binary attributes (e.g., Person-Non-person). For the following analysis, all of the collected subjects' responses (10 subjects for each counterfactual scene) are used to calculate human CSS maps, referred to as the human consensus.

\textbf{Sensitivity to Size:} Physical stimuli size acts as a primary bottom-up cue for visual attention in humans in early psychology research\citep{fitts1954information, castiello1990size}. However, deep neural networks exhibit an exaggerated, unnatural reliance on spatial dominance during feature extraction that is inconsistent with human perception, especially for object detection \citep{eckstein2017humans, hayes2021deep, wang2024poodle}. We hypothesize that this disproportionate reliance on size persists even within state-of-the-art VLMs, likely inherited from their foundational CNN or Vision Transformer architectures. We quantified the size of the object using the area of the masks.

\textbf{Sensitivity to Center:} Human free-viewing tends to fixate on the center of a scene \citep{tatler2007central}. Meanwhile, because large-scale training datasets (e.g., ImageNet
\citep{deng2009imagenet}, MS-COCO
\citep{lin2014microsoft}) are heavily saturated with centrally framed objects, deep learning models could inherit and amplify this spatial prior,  biasing their semantic interpretation to the center of the image regardless of contextual relevance. We used the negative of the distance between the center of the masks and the center of the image to measure the centeredness for each object.

\textbf{Sensitivity to Low-level Saliency:} Early theories of visual attention have shown that bottom-up, low-level image statistics, such as local contrasts in luminance, color, and orientation, automatically capture human gaze \citep{itti1998model}. To measure this low-level signal without introducing the confounding semantic variables present in modern neural networks, we quantified the low-level saliency of each ablated object using the Graph-Based Visual Saliency (GBVS) algorithm \citep{harel2006graph}. Specifically, the low-level visual saliency of each ablated object was quantified as the maximum GBVS activation value located within its corresponding segmentation mask.

\textbf{Sensitivity to Person/Non-person:}  Research in human eye movements demonstrates that observers consistently prioritize faces and human figures during free-viewing, independent of explicit task demands \citep{cerf2009faces, judd2009learning}. However, whether vision-language models intrinsically replicate this sensitivity during scene comprehension remains systematically underexplored. We utilized GPT-5.4 to classify each ablated object as either "person" or "non-person" based on its label during counterfactual stimuli generation.

\subsubsection{Divergence in Attribute Sensitivity}

\begin{figure}[ht]
    \centering
    
    \begin{subfigure}[b]{0.48\textwidth}
        \centering
        \includegraphics[width=\textwidth]{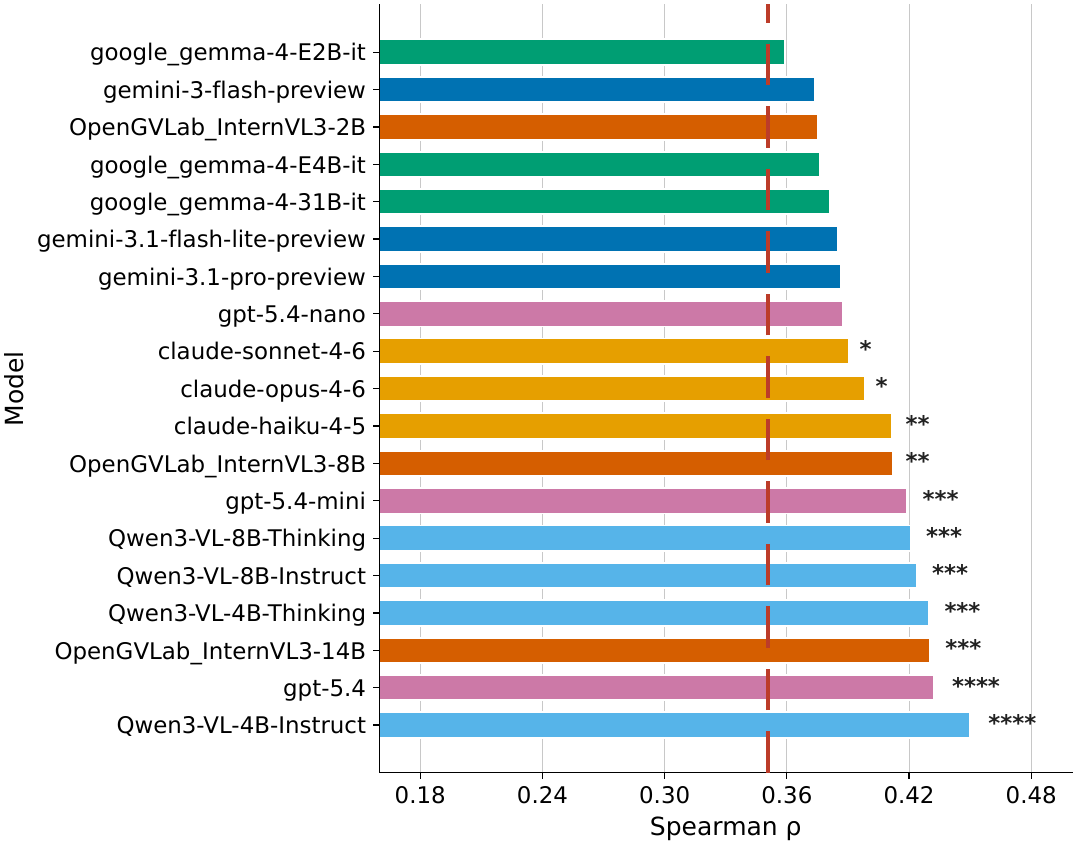}
        \caption{Sensitivity to Size}
        \label{fig:size_bias}
    \end{subfigure}
    \hfill 
    \begin{subfigure}[b]{0.48\textwidth}
        \centering
        \includegraphics[width=\textwidth]{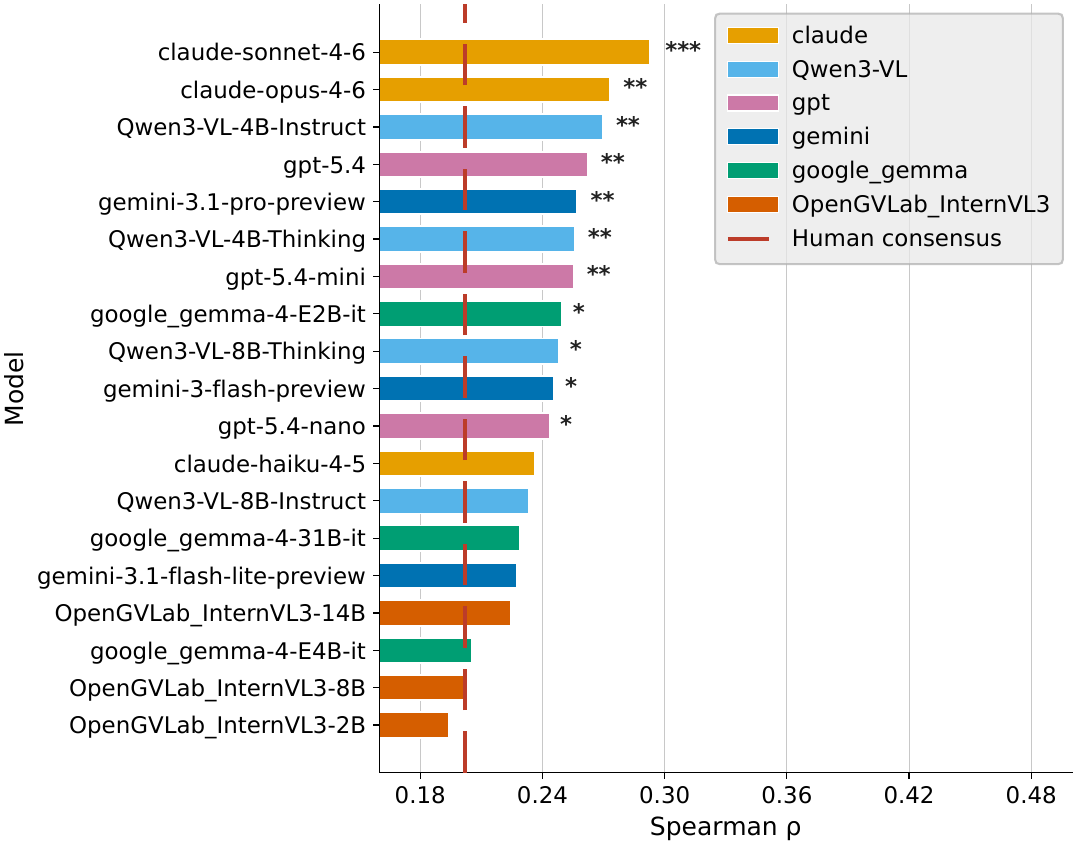}
        \caption{Sensitivity to Center}
        \label{fig:center_bias}
    \end{subfigure}
    
    \vspace{0.2cm} 
    
    \begin{subfigure}[b]{0.48\textwidth}
        \centering
        \includegraphics[width=\textwidth]{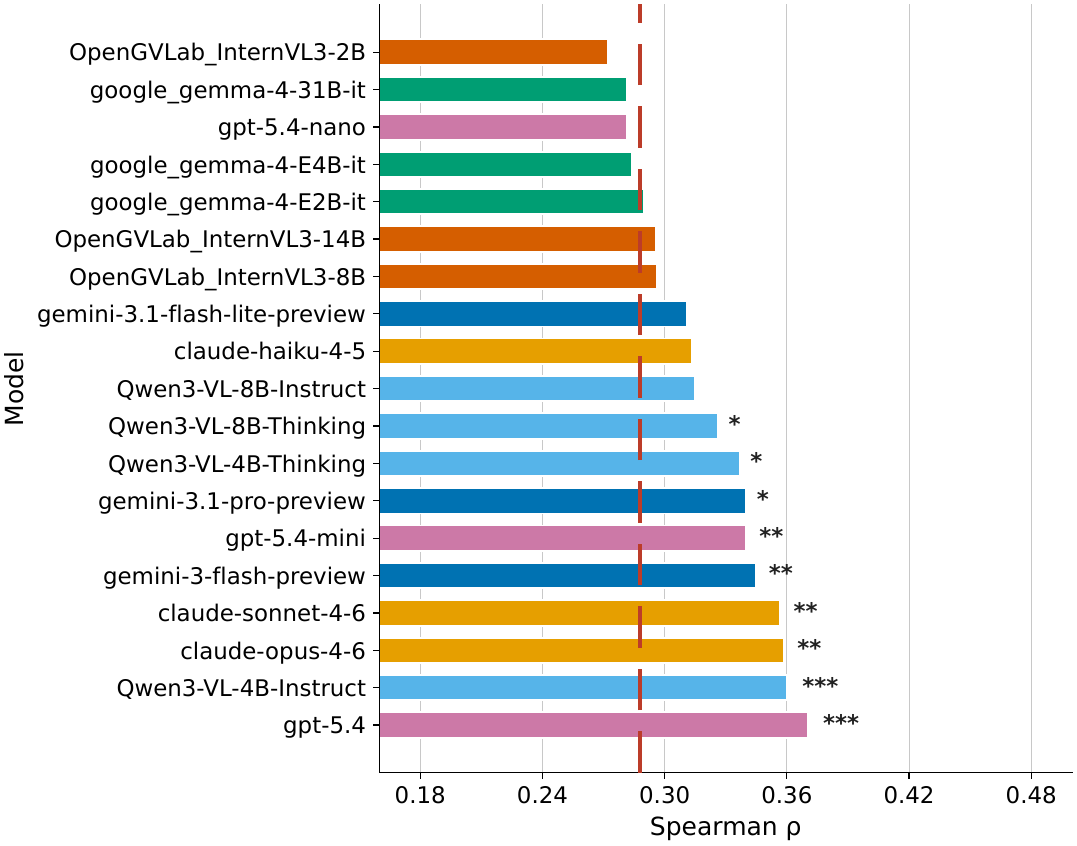}
        \caption{Sensitivity to Low-level Saliency}
        \label{fig:saliency_bias}
    \end{subfigure}
    \hfill
    \begin{subfigure}[b]{0.48\textwidth}
        \centering
        \includegraphics[width=\textwidth]{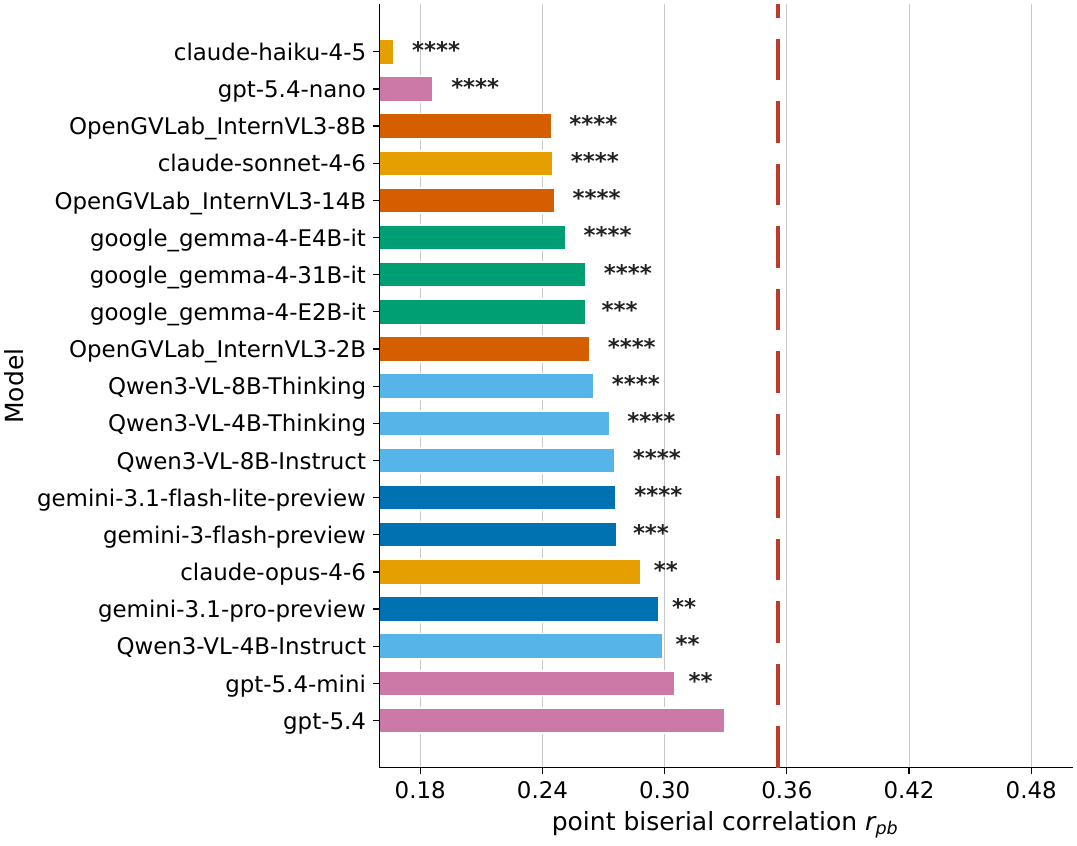}
        \caption{Sensitivity to Person-Non-person}
        \label{fig:person_bias}
    \end{subfigure}
    
    \caption{\textbf{Divergence of Attribute Sensitivity:} We computed the correlation between specific object attributes and the resulting CSS score upon ablation. Significance tests indicate the statistical deviation of each model's correlation coefficient from the human baseline ($* = p<.05$; $** = p<.01$; $*** = p<.001$; $**** = p<.0001$; derived from a one-tailed bootstrap test).}
    \label{fig:bias}
\end{figure}

To evaluate the four hypothesized drivers of semantic shift, we quantified the specific attribute sensitivity of all evaluated models and compared them directly against the human consensus baseline. As illustrated in Fig.\ref{fig:bias}, we measured the correlation between the target object attributes and the Counterfactual Semantic Saliency. To formally assess whether the models' reliance on these visual attributes significantly deviated from human perception, we conducted one-tailed bootstrap significance tests (n = 10,000) across the entire counterfactual image set.

Furthermore, to ensure the attribute sensitivity we observed from human consensus (dashed line in Fig.\ref{fig:bias}) was not an artifact of random statistical noise, we used permutation tests (n = 10,000) across all object attribute labels (Size, Center, Low-level Saliency, and Person-Non-person). Correlating the permuted data with the human consensus under the null hypothesis yielded a mean approximating zero with minimal variance ($std \approx 0.027$) for all 4 factors, confirming that the attribute sensitivity of humans identified here is statistically significant ($p<.0001$). The comparative analysis reveals a divergence in how VLM and humans prioritize information:

\begin{itemize}
    \item \textbf{Over-reliance on low-level attributes:} Deep learning models rely heavily on basic spatial and pixel-level features. Compared to human observers, models
    are significantly more sensitive to object size (Fig. \ref{fig:size_bias}), object centrality  (Fig. \ref{fig:center_bias}), and low-level saliency (Fig. \ref{fig:saliency_bias}). They are more frequently influenced by objects that are large, central, or highly contrasting, regardless of their actual semantic value. 
    \item \textbf{Insensitivity to the presence of people:} Conversely, models lack the specialized prioritization that humans apply to the scene. While the description from human viewers is heavily influenced by the presence of people in the scene, the evaluated models are less sensitive to whether people are present or absent in the scene. (Fig. \ref{fig:person_bias}). 
\end{itemize}

\subsection{Driving Factors for Semantic Shift Gap}

Having quantified the divergence of attribute sensitivity between VLMs and humans (Fig. \ref{fig:bias}), we need to determine which specific factors drive the observed disparity in critical object identification (Fig. \ref{fig:human_model_top1Acc}). To achieve this, we evaluated whether the magnitude of the sensitivity divergence ($\Delta r$) between models and humans for each specific attribute correlates with the overall performance gap in Top-1 Accuracy ($\Delta \text{Acc}$) between evaluated models and humans.

To isolate statistically significant driving factors from baseline random effects, we used permutation tests. We performed 10,000 permutations across the target attribute labels, recalculated the corresponding $\Delta r$ for the null distribution, and evaluated its correlation with $\Delta \text{Acc}$. As shown in Table \ref{tab:bias_gaps}, while the null distributions are centered near zero, the variance indicates a cautious interpretation of the correlations. Among the four hypotheses, Sensitivity to Size emerged as the dominant driving factor of the gap between humans and VLMs ($r = -0.6130$). The excessive reliance on object size explains the systematic model failure illustrated in Fig. \ref{fig:teaser}; because the white bird occupied a significantly larger pixel area than the fishing rod, the semantic shifts obscured the importance of the fishing rod, leading to misalignment with the human consensus.

\begin{table}[ht]
  \centering
  \caption{\textbf{Driving factors of the VLM-Human scene comprehension gap.} The correlation between the divergence in specific attribute sensitivity ($\Delta r$) and the overall gap in Top-1 Accuracy ($\Delta \text{Acc}$) across evaluated models. The null distribution metrics were obtained via permutation tests to isolate random baseline effects. The p-values here were obtained by one-tailed permutation tests}
    \label{tab:bias_gaps}
  \begin{tabular}{lcccc}
    \toprule 
    $\Delta r$ Type & \textbf{Corr. to $\Delta$Acc} & \textbf{Mean of Null Dist.} & \textbf{Std of Null Dist.} & \textbf{P-value} \\
    \midrule
    Size               & -0.6130 & 0.0023  & 0.3193 &  0.0211 \\
    Center             & -0.2296  & -0.0030 & 0.3189 & 0.2635 \\
    Low-level Saliency & -0.3746 & -0.0012 & 0.3155   &  0.1273\\
    Person-Non-person             & -0.1533 & -0.0014 & 0.3247  & 0.3398 \\
    \bottomrule
  \end{tabular}
\end{table}


\subsection{The Mismatch Between White-Box Attention and Black-Box Causality}\label{sec:white-box}
\begin{figure}[ht]
    \centering
    \includegraphics[width=\linewidth]{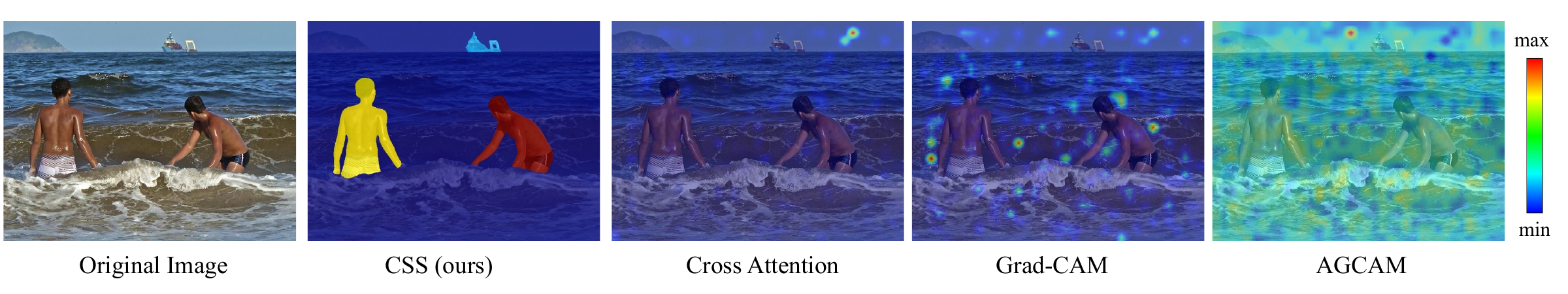}
    \caption{Comparison between CSS Maps and attention-based white-box maps }
    \label{fig:whitebox}
\end{figure}
Attention and activation maps have historically served as diagnostic tools for visualizing the features that contribute to a neural network's response. However, it remains an open question whether these white-box observational methods can yield interpretable, object-level importance in large VLMs comparable to our causal black-box framework. To compare CSS against traditional interpretability techniques, we applied three widely adopted methods, raw Attention maps, Grad-CAM\citep{selvaraju2017grad}, and Attention-Guided CAM (AGCAM)\citep{leem2024attention}, to an open-source VLM (Qwen3-VL-4B-Instruct).

Because VLM text generation is autoregressive, these white-box methods yield distinct spatial saliency maps (not to be confused with low-level saliency maps such as GBVS) for every generated token. To capture the global influence of the input pixels on the entire generated description, we aggregate all the saliency maps generated for each token through a max function. Finally, to enable a direct comparison with our object-level CSS metric, the white-box saliency of each visual element is formally quantified as the maximum activation value located within its corresponding segmentation mask.

Qualitative comparisons between these raw white-box saliency maps (before object masking) and CSS are illustrated in Fig. \ref{fig:whitebox}. Notably, the white-box maps derived from token-image activations severely lack human interpretability; they are frequently either incomprehensibly cluttered or overly sparse, and often focus on semantically irrelevant spatial regions. Consequently, the correlation between these traditional object-level saliency scores and the CSS metric is low and negative (mean Kendall's $\tau$ across all the images with the CSS: raw attention: -0.3352, GradCAM: -0.2998, AGCAM: -0.2420). This mismatch reveals the limitations of passive interpretability techniques when applied to the complex architectures of modern VLMs. More comparisons between our methods and white-box saliency maps are provided in the Appendix.

\subsection{Vision-Only Models' Observational Saliency Fails to Capture CSS}

Salient Object Detection (SOD) models \citep{qin2020u2,liu2021visual} are trained on large-scale datasets of human-annotated salient object masks, providing a vision-only estimate of which objects humans perceive as critical in a scene. However, similar to the white-box methods discussed in Sec.~\ref{sec:white-box}, SOD remains an observational approach: it infers object importance passively from image appearance without intervening on the scene. We therefore ask whether such a vision-only method can predict the object importance revealed by CSS.

Here, we evaluated two commonly used SOD models, $U^2$-Net \citep{qin2020u2} and Visual Saliency Transformer (VST) \citep{liu2021visual}, on the same 307 original images used in our evaluation. Following a similar procedure to that used for the attention-based white-box maps in Sec.\ref{sec:white-box}, we used the maximum and mean values of the $U^2$-Net /VST saliency map within the corresponding segmentation mask in CSS to convert the pixel-level saliency maps into object-level saliency maps. Those SOD maps are treated the same way as the VLMs' CSS maps, and we repeat the same evaluation reported in Fig.\ref{fig:overall_eval}. The performance of U2Net and VST using both maximum and mean values is shown in Table.\ref{tab:SOD}. The performance of Top1-Acc from both SOD models falls below all the VLMs we evaluated, indicating that vision-only approaches, even when trained on human-annotated data, struggle to capture the same object-level importance as CSS.

\begin{table}[ht]
  \caption{Performance of Salient Object Detection on predicting the critical object and rank order of object importance. gemma-4-E4B represents the VLM with the worst Top-1 Acc performance. }
  \label{tab:SOD}
  \centering
  \begin{tabular}{lcc}
    \toprule
    Baseline & Mean $\tau$ $\uparrow$ & Top1-Acc $\uparrow$ \\
    \midrule
    VST (max) \citep{liu2021visual}       & 0.378 & 0.414 \\
    VST (mean)  \citep{liu2021visual}    & 0.258          & 0.345 \\
    $U^2$-Net (max) \citep{qin2020u2}    & 0.353          & 0.476 \\
    $U^2$-Net(mean) \citep{qin2020u2}   & 0.229          & 0.384 \\
    gemma-4-E4B    &   0.401             &     0.573           \\
    \bottomrule
  \end{tabular}
\end{table}

\section{Discussion}
\textbf{Data Distribution and Inductive Bias}: The over-reliance on low-level attributes from the models we discovered is likely to
originate from the "object-centered" nature of foundational vision datasets. Most vision encoders are pretrained with images in which the primary subject is centrally framed and spatially dominant \citep{taesiri2023imagenet}, leading the encoder to mistake size for importance. Consequently, smaller or peripheral objects, which may be semantically indispensable to a human viewer, show only a minor influence on the model’s internal representation. This dependence is further compounded during the vision-language alignment \citep{geirhos2020shortcut}. Most image-caption pairs sourced from the Internet typically focus on spatially dominant entities \citep{gordon2013reporting, misra2016seeing}. As a result, even if the vision encoder captures a complete scene representation, the language output remains more sensitive to object size regardless of its actual semantic importance.


\textbf{Superhuman Inference and Reasoning Gap:} The "superhuman" capabilities of modern VLMs, specifically their advanced reasoning, can also exacerbate the human-AI mismatch. An example is illustrated in Fig.\ref{fig:teaser}: the ablation of the fishing rod fundamentally shifted the human interpretation of the scene (e.g., "fishing equipment" -> "a bunch of trash"). In contrast, models were often able to infer the ablated context from surrounding cues or, in some cases, exhibited hallucinatory behavior by describing the ablated object as if it were still present \citep{rohrbach2018object, biten2022let, li2023evaluating}.


\textbf{Limitation:} While this work characterizes how the alignment gap between models and humans manifests in scene description, the underlying causes of this divergence remain unexplored. For example, the models' insensitivity to the presence or absence of a person in the scene could arise because they hallucinate people in scenes where none are present, or because they mention people less often than human observers do. Either case could produce the observed divergence. Identifying which factors are responsible will be crucial for developing vision-language models that are better aligned with human perception.

\section{Conclusion}

We introduced Counterfactual Semantic Saliency (CSS), a novel black-box framework designed to quantify the importance of objects in scene comprehension. By benchmarking modern VLMs against a large-scale human psychophysics baseline, we uncovered a pervasive alignment gap in scene comprehension. We demonstrate that while human observers naturally rely on the presence of people in the scene, modern VLMs default to an over-reliance on low-level visual features (object size, centrality, saliency).

\section{Acknowledgment}
Research was sponsored by the U.S. Army Research Office and accomplished under cooperative agreement W911NF-19-2-0026 for the Institute for Collaborative Biotechnologies.

\newpage
\bibliographystyle{abbrv}
\bibliography{ref}

\newpage
\appendix

\section{Technical Details of Counterfactual Semantic Saliency}

This section shows the prompts and hyperparameters employed in the Counterfactual Semantic Saliency (CSS) pipeline.

For the object name generation stage, we utilized GPT-4o\citep{achiam2023gpt}. For each factual scene, we sampled two independent responses using the prompt detailed below to maximize object coverage. Empirical testing indicated that sampling more than twice yielded negligible marginal gains in discovering distinct, new objects. The sampling temperature was set to 1.0, with all other hyperparameters at their default values.

Crucially, the objective of this approach is not the exhaustive enumeration of every entity in the scene. Rather, it is to generate a controlled counterfactual environment for robust human and model evaluation. Consequently, guaranteeing absolute recall of all objects is not strictly necessary for our methodology.

\begin{tcolorbox}[
  title=Prompts for Object List Generation, 
  colback=gray!5!white,                    
  colframe=gray!75!black,                  
  colbacktitle=gray!20!white,              
  coltitle=black,                          
  fonttitle=\bfseries,                     
  breakable,                               
  enhanced                                 
]

List the names of all objects in the scene, do NOT mention the objects that would be considered as background (like sky, sea, ground), each separated by a comma. The name of the object should be short. The format MUST looks like: 'red car, blue car, man in black'. If you cannot identify any object, please write 'None'.

\end{tcolorbox}

\begin{tcolorbox}[
  title=Prompts for Object List Merging, 
  colback=gray!5!white,                    
  colframe=gray!75!black,                  
  colbacktitle=gray!20!white,              
  coltitle=black,                          
  fonttitle=\bfseries,                     
  breakable,                               
  enhanced                                 
]

I have \{len(caption\_list)\} lists of object names from the same image. Please merge them into one list, removing duplicates.

Objects that refer to the same thing should be considered duplicates even if worded slightly differently (e.g., "man in black" and "person in black" are duplicates).
Keep the most descriptive version when there are duplicates. If the object name contains two entries (e.g., "man with umbrella"), split them into two separate objects ("man with umbrella" and "umbrella").

\{list\_lines\}

Please return the merged list as a comma-separated string in the same format as the input. For example: 'red car, blue car, man in black'.
If all lists contain "None" or no valid objects, return "None". You should ONLY return the list of object names, NO other text.

\end{tcolorbox}

Object removal was executed using the Nano Banana 2 model via the Gemini API. To guide the generative inpainting process and ensure the semantic integrity of the underlying scene, we overlaid the factual scene with a transparent red mask targeting the specified object. Such masks are obtained through the SOTA text-prompted segmentation model SAM3 \citep{carion2025sam}, see the main content for how the masks are processed before being applied to the Object Removal Stage. The segmentation process is computed by using a single GeForce RTX 4090 GPU.

\begin{tcolorbox}[
  title=Prompts for Object Removal, 
  colback=gray!5!white,                    
  colframe=gray!75!black,                  
  colbacktitle=gray!20!white,              
  coltitle=black,                          
  fonttitle=\bfseries,                     
  breakable,                               
  enhanced                                 
]

Remove the object covered with red masks; ONLY remove the object itself.

\end{tcolorbox}

During the VLM evaluation using the Counterfactual Semantic Saliency, we applied the prompt detailed below to all evaluated models for every image. To account for generation variability, we sampled five independent descriptions per image. An ablation study comparing these results against deterministic sampling is presented in the subsequent section.

For all models, the maximum generation length was set to 16,384 tokens to ensure the capture of complete descriptions. For such reasoning models, only the final generated scene description was extracted for evaluation. All other sampling parameters were maintained at their default settings, whether the models were accessed via Hugging Face or API call. For open-source model evaluations, all are executed on a single NVIDIA RTX PRO 6000 GPU.

\begin{tcolorbox}[
  title=Prompts for Scene Description, 
  colback=gray!5!white,                    
  colframe=gray!75!black,                  
  colbacktitle=gray!20!white,              
  coltitle=black,                          
  fonttitle=\bfseries,                     
  breakable,                               
  enhanced                                 
]

Make your best guess of what might be happening in this scene in one sentence. Avoid mentioning objects that do not aid in understanding the context of the scene.

\end{tcolorbox}

\section{Data Quality Validation Protocols}\label{sec:data_qaulity}

To ensure the high fidelity of our counterfactual scene generation, we recruited 14 human annotators via Prolific to verify the successful removal of target objects. Annotators were presented with side-by-side image pairs: the original factual scene is covered with a red mask highlighting the target object, and the generated counterfactual scene where the object was intended to be ablated. 

To maintain annotation quality, the generated images were divided into 7 distinct batches, capping each batch at fewer than 300 trials. Two independent annotators were assigned to evaluate each batch. Their specific instructions for identifying successful object ablation are detailed below:

\begin{tcolorbox}[
  title=Instructions for Data Quality Validation, 
  colback=gray!5!white,                    
  colframe=gray!75!black,                  
  colbacktitle=gray!20!white,              
  coltitle=black,                          
  fonttitle=\bfseries,                     
  breakable,                               
  enhanced                                 
]
In this experiment, each time you will be given a pair of images. The image on the right is an image with a red mask on a particular object. The image on the left is the image where that object is removed.

Your task is to identify whether the object with a red mask is removed in the left image. The answer is always "Yes" or "No":

When the object with red masks is successfully removed, select “Yes”.

When the object with red masks is not successfully removed, select “No”. 
\end{tcolorbox}

\begin{figure}[ht]
    \centering
    \includegraphics[width=\linewidth]{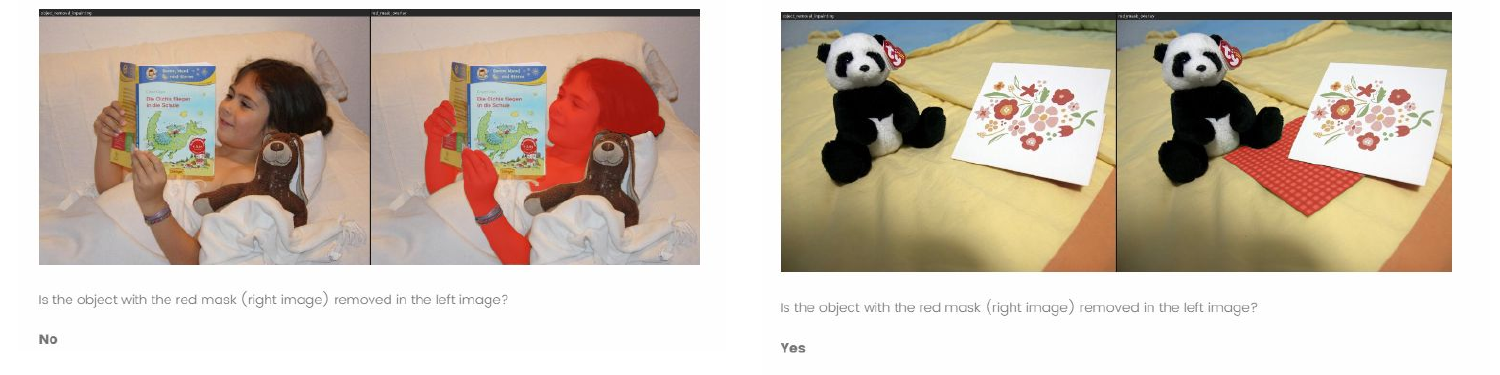}
    \caption{Examples provided for annotators in data quality validation process.}
    \label{fig:appendix_quality}
\end{figure}

Prior to the main task, annotators were provided with standardized examples of successful (``Yes'') and failed (``No'') object removals as illustrated in Fig.\ref{fig:appendix_quality}. During the experiment, stimuli were presented using this exact standardized layout.

Upon collection of the crowdsourced annotations, any counterfactual image flagged with a ``No'' by either of the two assigned annotators underwent a secondary manual review by the researchers. Images definitively exhibiting failed ablations were discarded. Furthermore, if the discarding of failed generations caused a factual scene to fall below a minimum threshold of three valid counterfactual variants, the entire factual scene and its remaining variants were removed from the dataset. Following this filtering protocol, the finalized dataset comprises 307 unique factual scenes and 1,306 high-fidelity counterfactual variants.

\section{Online Psychophysics Details}

This section contains the explicit instructions for our large-scale online data collection. A comprehensive overview of the core psychophysical experimental design is provided in the main text.

Participants were recruited via the Prolific crowdsourcing platform. To ensure high-quality textual descriptions, eligibility was strictly restricted to native or highly fluent English speakers who maintained a historical task approval rate of $\geq 98\%$. To ensure strict methodological parity between the human baseline and the  models, the primary task instruction presented to human participants, \textit{“Make your best guess of what might be happening in this scene in one sentence. Avoid mentioning objects that do not aid in understanding the context of the scene.”}, was identical to the text prompt utilized for the evaluated Vision-Language Models.

\begin{tcolorbox}[
  title=Instructions for Online Psychophysics,
  colback=gray!5!white,                    
  colframe=gray!75!black,                  
  colbacktitle=gray!20!white,              
  coltitle=black,                          
  fonttitle=\bfseries,                     
  breakable,                               
  enhanced                                 
]
Your task will be to look at the image and carefully provide a description. 

\textbf{Make your best guess of what might be happening in this scene in one sentence. Avoid mentioning objects that do not aid in understanding the context of the scene}.

We expect the following for the descriptions:

1. The description should explain what is happening in the scene.

2. The descriptions need to be concise.

The first three images have been provided with descriptions as an example. 
Please carefully review the examples, as they will give you an idea of the kind of images you will see in the survey and the kind of descriptions we expect.

You need to satisfy these requirements to participate:

1. You MUST be a Native English speaker. This means that you were raised speaking English

2. You MUST carefully look at the example shown and provide descriptions as suggested.

3. You MUST thoroughly review each image and provide a meaningful and grammatically correct description.

4. Please ensure to open this link on a laptop or Desktop.
\end{tcolorbox}

\begin{figure}[ht]
    \centering
    \includegraphics[width=0.85\linewidth]{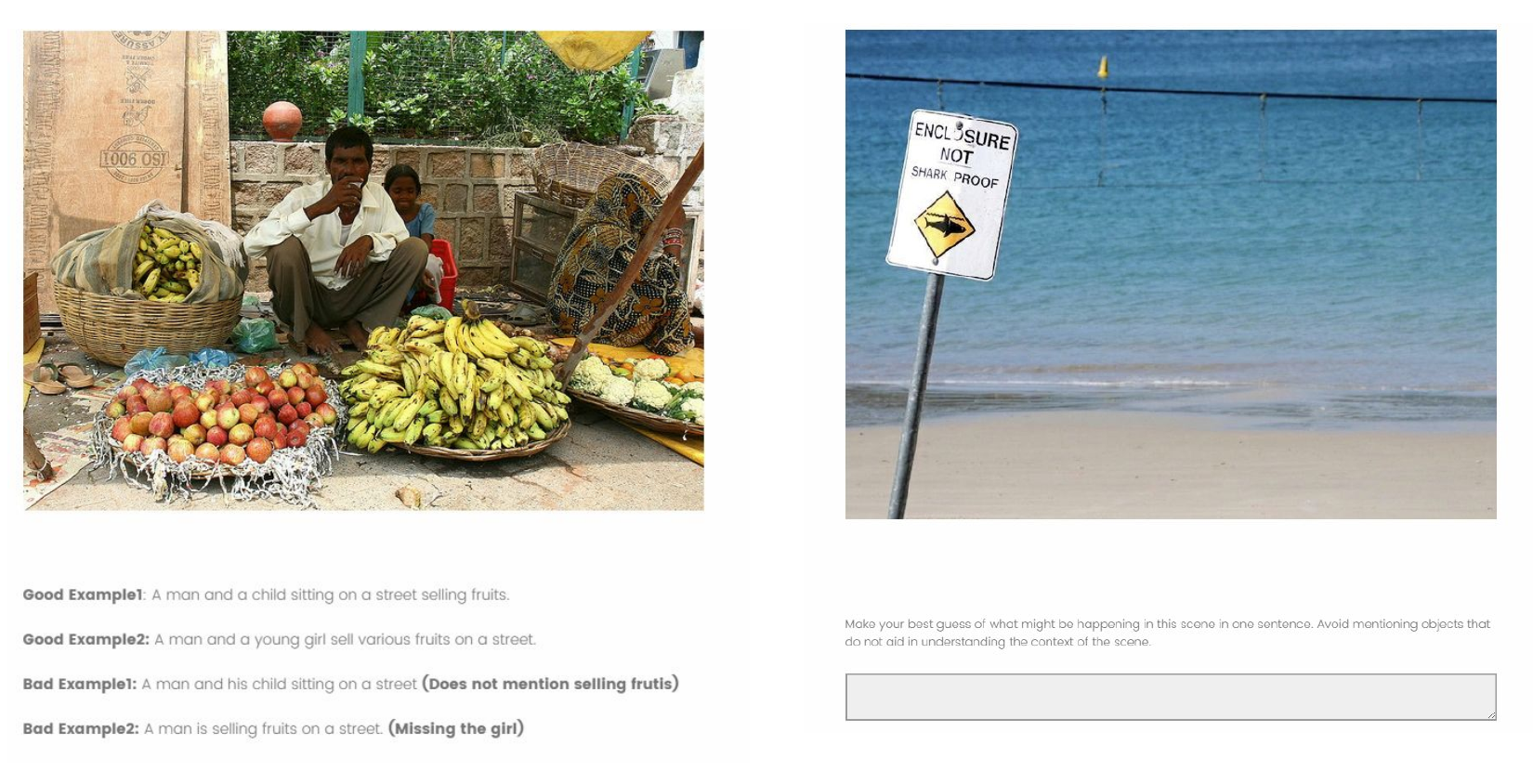}
    \caption{\textbf{Examples of Online Psychophysics}: Left: One of the examples provided to subjects as calibrations. Right: One of the trails within the experiment.}
    \label{fig:online_study_example}
\end{figure}

To standardize the verbosity of human responses, participants were required to review examples of acceptable textual descriptions before beginning the main experimental trials (Fig. \ref{fig:online_study_example}). Importantly, the visual stimuli utilized for these calibration examples were strictly disjoint from the main dataset used for the formal human study and the VLM evaluations. Subjects can have both factual and counterfactual scenes in their psychophysics, while they are not informed that some of the scenes are generated object-ablated scenes. 

Following data collection, participants identified as potential automated bots or those exhibiting consistently low-effort responses were completely excluded from the dataset. Furthermore, responses were discarded if their mean semantic similarity to other human responses for the same image was below 0.4, as measured by sentence embedding \cite{zhang2024jasper}.

\section{Ablations: Deterministic vs. Stochastic Decoding}

As detailed in the main text, our primary evaluation employs stochastic sampling during VLM generation to capture the variance and distribution inherent in human scene perception. In this section, we present an ablation study investigating the impact of restricting models to a single deterministic output. Specifically, we re-evaluate the seven open-source models featured in our primary results (e.g., Qwen3-VL, Gemma4, and InternVL3) by generating their textual descriptions using greedy decoding. We then quantify their semantic alignment with the human cognitive baseline using Top-1 Accuracy and Kendall's $\tau$.

\begin{figure}[htbp]
    \centering
    \begin{subfigure}[b]{0.48\textwidth}
        \centering
        \includegraphics[width=\textwidth]{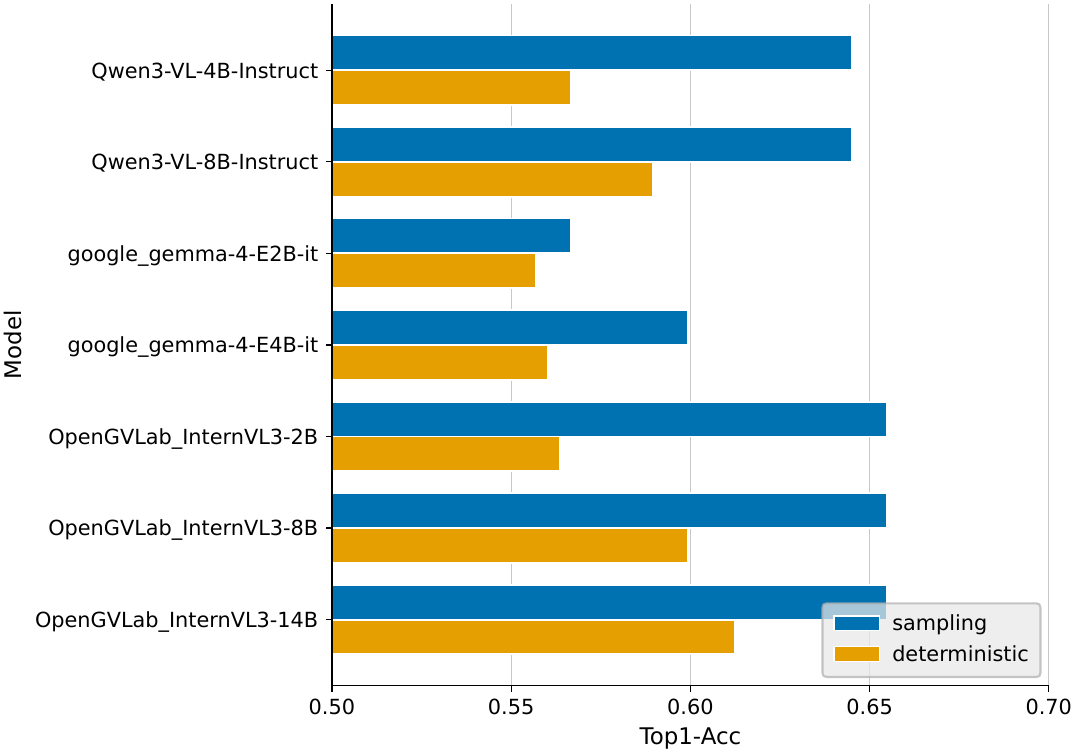}
        \caption{Top-1 Accuracy at predicting the critical object.}
    \end{subfigure}
    \hfill 
    \begin{subfigure}[b]{0.48\textwidth}
        \centering
        \includegraphics[width=\textwidth]{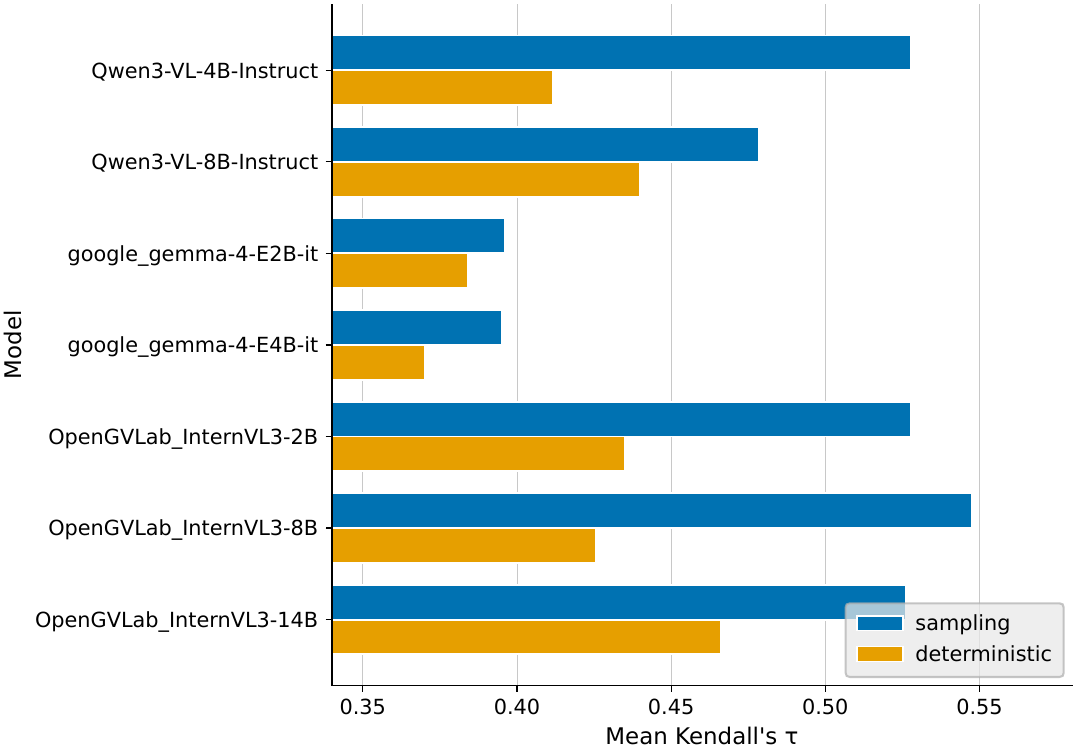}
        \caption{Mean Kendall's $\tau$  between models and humans.}
    \end{subfigure}
    
    \caption{Comparison between deterministic output and stochastic sampling. \textit{Sampling} results are the same results in the main paper.}
    \label{fig:deterministic}
\end{figure}

As illustrated in Fig.\ref{fig:deterministic}, models evaluated under deterministic decoding exhibit a significant performance drop compared to their sampling counterparts. This drop in human-AI alignment indicates that introducing variability via stochastic sampling is necessary for models to accurately approximate the true distribution of human responses.

\section{Robustness across Inpainting Methods}
\label{app:inpainting}

Because CSS relies on generative models to construct counterfactual scenes, the inpainting method itself could be a confound. We consider two such confounds. First, the inpainting model reconstructs the masked region according to its own prior, so the observed semantic shift might partially reflect properties of the inpainting model rather than the removal of the target object. Second, our analysis assumes that object removal is the only meaningful difference between a factual scene and its counterfactual variant. A generative model such as Nano Banana 2 may also alter content outside the target mask, and such changes could correlate with object size. In this section, we address each concern in turn and show that neither accounts for the model--human gap or the attribute sensitivity reported in the main text.

\subsection{Replacing the Inpainting Model}
\label{app:inpainting_mat}

To test whether our findings depend on the prior of a specific inpainting model, we regenerated all counterfactual stimuli using the Mask-Aware Transformer (MAT)~\citep{li2022mat}, an inpainting model based on GAN whose generative prior differs fundamentally from that of the diffusion-based Nano Banana 2. Although MAT may introduce visible artifacts in some images, its distinct prior makes it a useful test of whether the scene comprehension gap and the attribute sensitivity are specific to a particular inpainting model. Using the same segmentation masks and preprocessing pipeline as in the main experiments, we removed the same target objects and obtained a new set of 1,306 counterfactual scenes. We then repeated the full evaluation of all 19 VLMs, including the Top-1 Accuracy analysis (Fig.~\ref{fig:overall_eval}a) and the four attribute sensitivity analyses (Fig.~\ref{fig:bias}).

\begin{figure}[ht]
  \centering
  \includegraphics[width=0.65\linewidth]{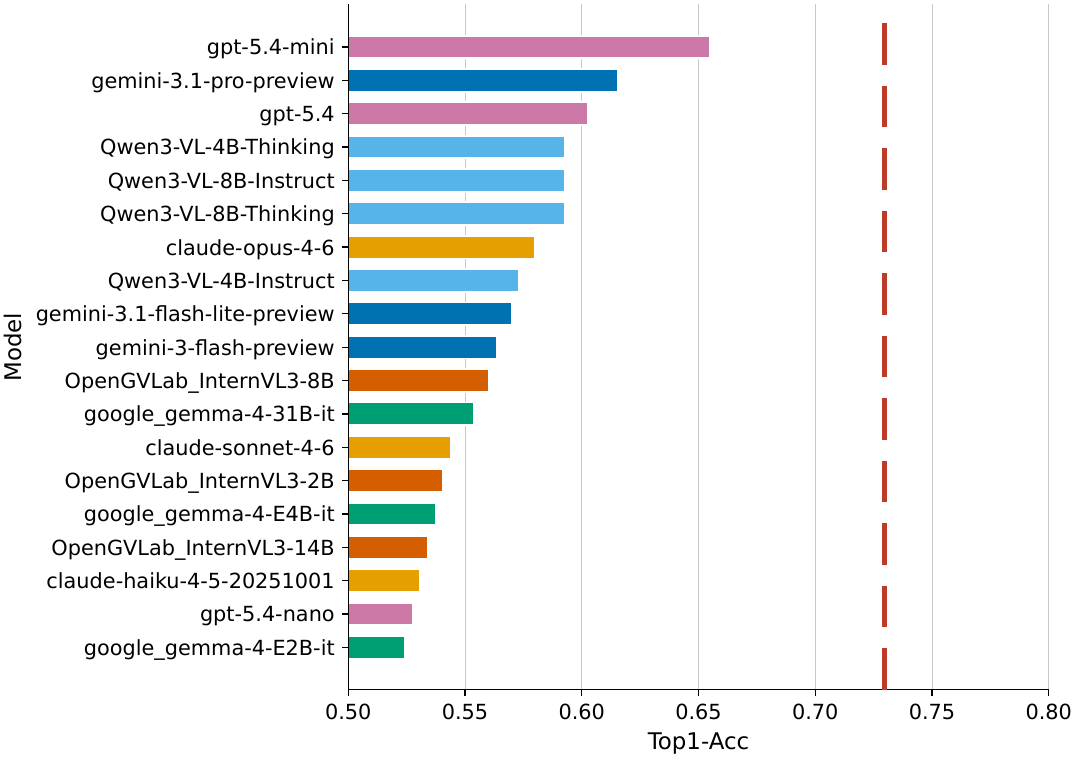}
  \caption{Top-1 Accuracy at predicting the critical object when counterfactual scenes are generated with MAT~\citep{li2022mat}. The dashed line denotes human--human consistency. All gaps are significant (one-tailed bootstrap test, $p < .01$).}
  \label{fig:mat_acc}
\end{figure}

\begin{figure}[ht]
  \centering
  \begin{subfigure}[b]{0.48\linewidth}
    \centering
    \includegraphics[width=\linewidth]{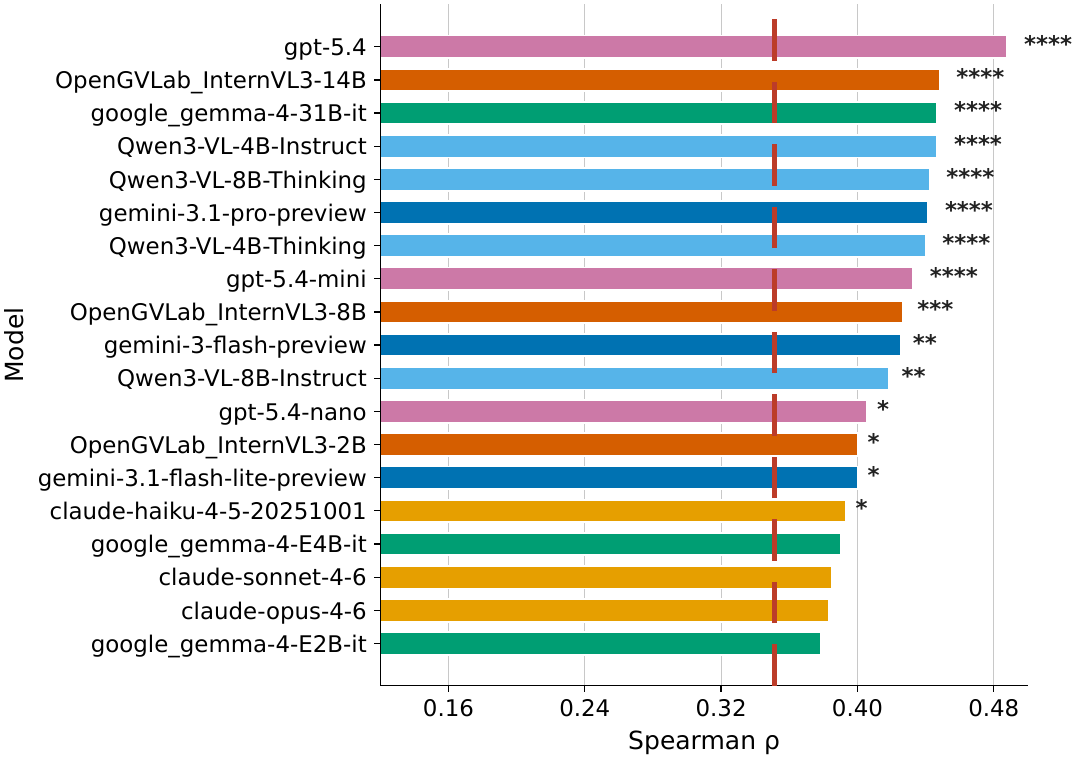}
    \caption{Sensitivity to Size}
    \label{fig:mat_bias_size}
  \end{subfigure}
  \hfill
  \begin{subfigure}[b]{0.48\linewidth}
    \centering
    \includegraphics[width=\linewidth]{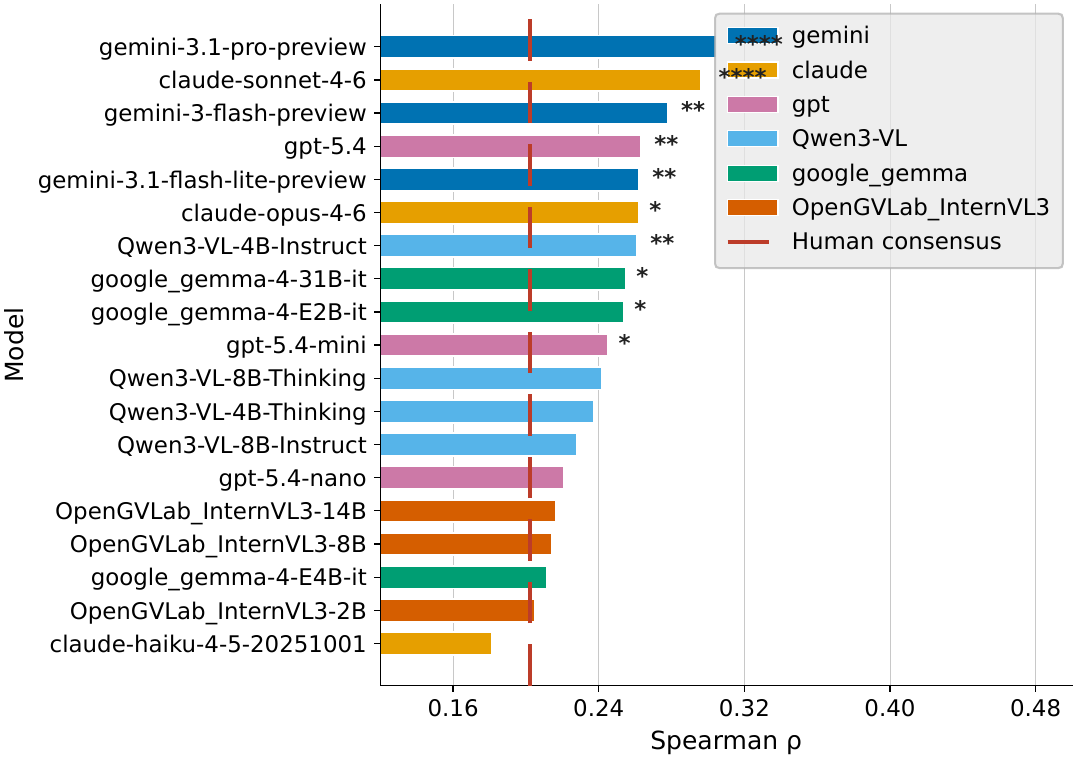}
    \caption{Sensitivity to Center}
    \label{fig:mat_bias_center}
  \end{subfigure}

  \vspace{0.5em}

  \begin{subfigure}[b]{0.48\linewidth}
    \centering
    \includegraphics[width=\linewidth]{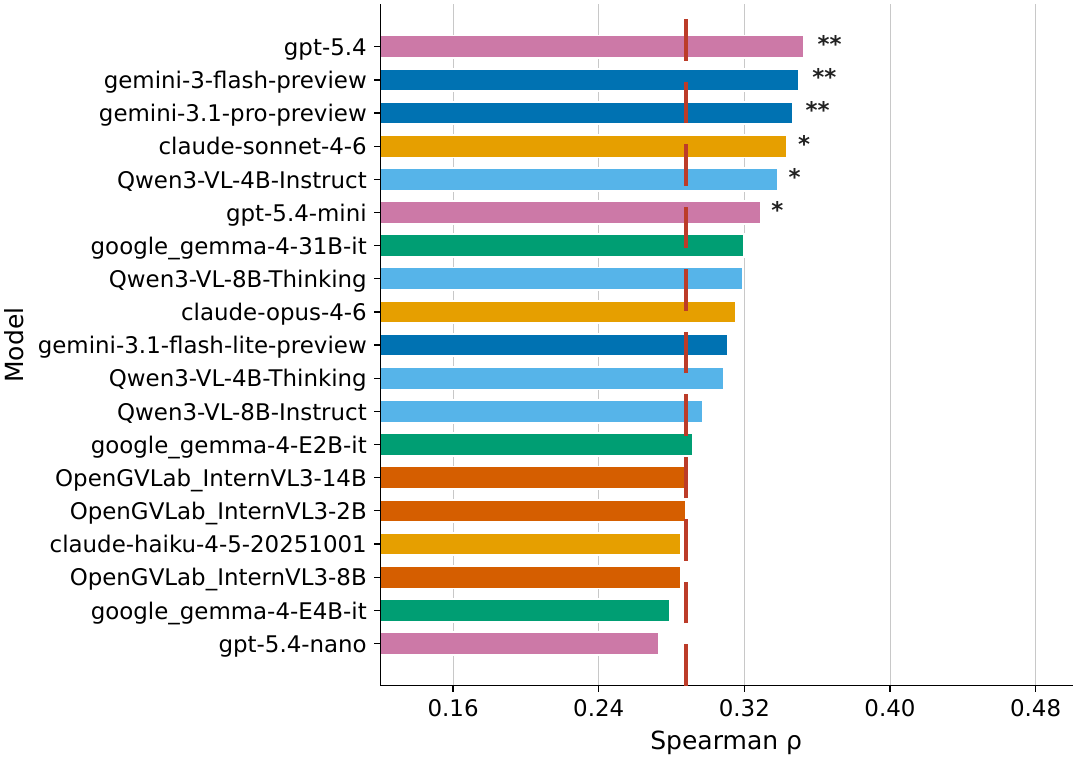}
    \caption{Sensitivity to Low-level Saliency}
    \label{fig:mat_bias_lowlevel}
  \end{subfigure}
  \hfill
  \begin{subfigure}[b]{0.48\linewidth}
    \centering
    \includegraphics[width=\linewidth]{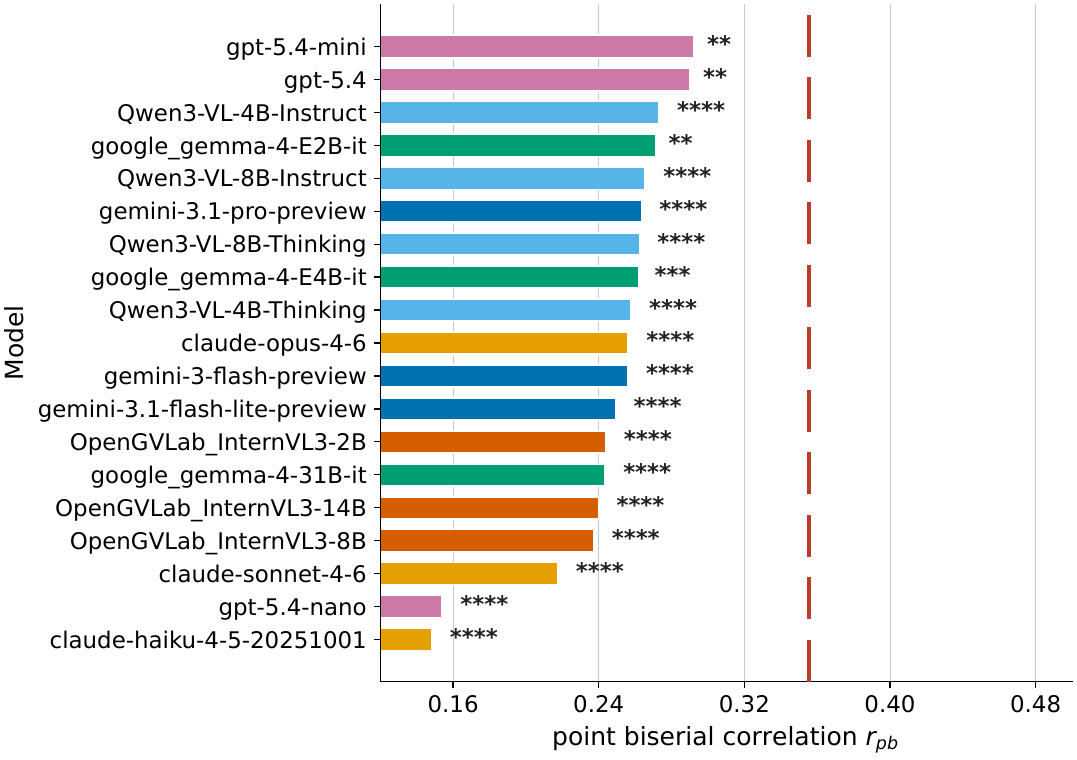}
    \caption{Sensitivity to Person-Non-person}
    \label{fig:mat_bias_person}
  \end{subfigure}

  \caption{Attribute Sensitivity of VLMs when counterfactual scenes are generated with MAT~\citep{li2022mat}.}
  \label{fig:mat_bias}
\end{figure}

As shown in Fig.~\ref{fig:mat_acc}, the systematic gap between human--human consistency and model Top-1 Accuracy persists and remains statistically significant under MAT inpainting. Likewise, the divergences in all four attribute sensitivities are preserved (Fig.~\ref{fig:mat_bias}): models continue to rely more on object size, object in the center, and Low-level Saliency, while being less sensitive to whether the object is a person. These results indicate that the alignment gaps are not artifacts of the Nano Banana 2 prior.

\subsection{Changes Outside the Object Mask}
\label{app:inpainting_outside}

Generative editing models may alter image content beyond the target region. For example, in Fig.~\ref{fig:method}, removing the TV also removes the lighting effect of the pink LED on the surrounding wall. Although our human validation (Appendix~\ref{sec:data_qaulity}) ensures that the target objects are successfully removed, it does not rule out the possibility that such out-of-mask changes correlate with object size and thereby contribute to the observed over-reliance to size.

\paragraph{Measuring out-of-mask changes.}
To quantify changes outside the target region without confounding them with mask size, we compute a modified Peak Signal-to-Noise Ratio (PSNR) that excludes the masked region. Let $I$ and $I'$ denote the factual and counterfactual images, and let $\tilde{M}$ denote the target mask dilated with a $24 \times 24$ kernel, identical to the dilation applied when preparing the inputs to the inpainting model. The mean squared error is computed only over the pixels outside the dilated mask, and we refer to the resulting metric as $\mathrm{PSNR}_{\text{out}}$. Across all 1,306 counterfactual scenes generated by Nano Banana 2, PSNRout is only weakly correlated with mask size (Pearson r=-0.166), suggesting that the size of the removed object has a minor influence on changes to the rest of the scene.


\paragraph{Out-of-mask changes do not explain the Sensitivity to Size.}
If the Sensitivity to Size reported in Fig.~\ref{fig:bias}a were driven by out-of-mask changes, these changes should affect models and humans differently. In that case, replacing object size with $\mathrm{PSNR}_{\text{out}}$ in the sensitivity analysis should reproduce a similarly systematic and significant model--human gap. We therefore repeated the analysis of Fig.~\ref{fig:bias}, computing the Spearman correlation between $\mathrm{PSNR}_{\text{out}}$ and CSS for each model and for humans (human $\rho = 0.210$).

\begin{figure}[ht]
  \centering
  \includegraphics[width=0.6\linewidth]{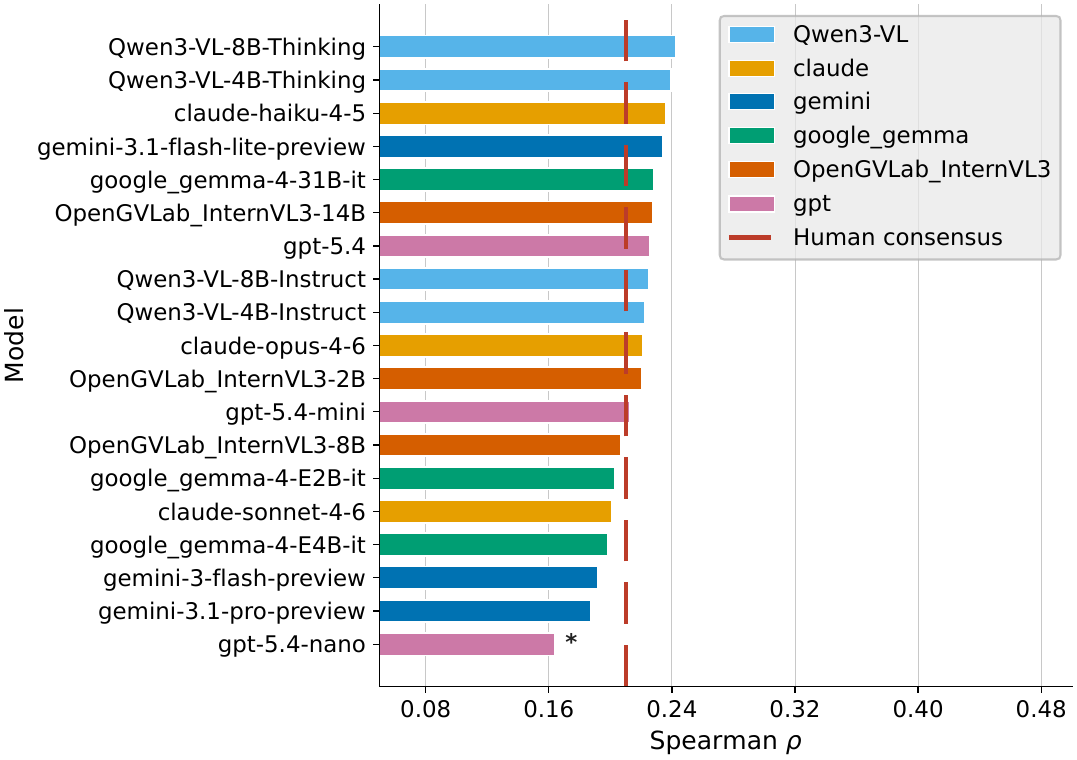}
  \caption{Spearman correlation between the modified PSNR outside the object mask ($\mathrm{PSNR}_{\text{out}}$) and CSS for each evaluated model.}
  \label{fig:psnr_bias}
\end{figure}

As shown in Fig.~\ref{fig:psnr_bias}, the pattern differs markedly from that of the Sensitivity to Size. With the exception of gpt-5.4-nano, no model differs significantly from humans in its sensitivity to out-of-mask changes. Moreover, gpt-5.4-nano shows a significantly \emph{weaker} correlation than humans in this analysis, whereas it shows a significantly \emph{stronger}  Sensitivity to Size than humans in Fig.~\ref{fig:bias}a. Thus, out-of-mask changes affect models and humans similarly and cannot account for the divergence of sensitivity to size observed in the main text.

\section{Robustness across Sentence Embedding Models}
\label{app:embedding}

Here we show that the divergence between model and human through CSS is robust to the change of sentence embedding model. We replaced the embedding model used in the main experiments, Jasper (\texttt{jasper\_en\_vision\_language\_v1})~\citep{zhang2024jasper}, with two alternatives: the open-source Qwen3-Embedding-0.6B~\citep{zhang2025qwen3} and the closed-source gemini-embedding-001~\citep{lee2025gemini}. We examined the robustness of our results at two levels: the CSS scores themselves (Sec.~\ref{app:embedding_css}) and the model--human comparisons (Sec.~\ref{app:embedding_gap}).

\subsection{Consistency of CSS Scores}
\label{app:embedding_css}

For each evaluated VLM and for the human baseline, we recomputed the CSS score of every counterfactual scene (1,306 in total) using each of the three embedding models. We then measured the Spearman rank correlation between the CSS scores obtained with each pair of embedding models to check if changing the embedding model has a non-negligible impact on the CSS score. As shown in Table~\ref{tab:embedding_css}, the rank order of CSS scores changes little across embedding models. For all VLMs and for humans, the Spearman correlations between embedding models are consistently close to 0.95, indicating that CSS itself is robust to the choice of sentence embedding model.

\begin{table}[ht]
  \caption{Spearman rank correlation of CSS scores computed with different sentence embedding models across all 1,306 counterfactual scenes. J: Jasper (used in the main experiments); Q: Qwen3-Embedding-0.6B; G: gemini-embedding-001.}
  \label{tab:embedding_css}
  \centering
  \begin{tabular}{lccc}
    \toprule
    Source & J vs.\ Q & J vs.\ G & Q vs.\ G \\
    \midrule
    Human                           & 0.944 & 0.945 & 0.933 \\
    \midrule
    claude-haiku-4-5                & 0.965 & 0.965 & 0.956 \\
    claude-opus-4-6                 & 0.957 & 0.957 & 0.949 \\
    claude-sonnet-4-6               & 0.963 & 0.954 & 0.942 \\
    gemini-3-flash-preview          & 0.958 & 0.962 & 0.956 \\
    gemini-3.1-flash-lite-preview   & 0.961 & 0.963 & 0.958 \\
    gemini-3.1-pro-preview          & 0.966 & 0.962 & 0.957 \\
    gemma-4-31B-it                  & 0.968 & 0.975 & 0.964 \\
    gemma-4-E2B-it                  & 0.968 & 0.975 & 0.965 \\
    gemma-4-E4B-it                  & 0.958 & 0.969 & 0.951 \\
    gpt-5.4                         & 0.966 & 0.973 & 0.963 \\
    gpt-5.4-mini                    & 0.963 & 0.970 & 0.961 \\
    gpt-5.4-nano                    & 0.952 & 0.958 & 0.953 \\
    InternVL3-2B                    & 0.938 & 0.938 & 0.926 \\
    InternVL3-8B                    & 0.946 & 0.958 & 0.940 \\
    InternVL3-14B                   & 0.946 & 0.958 & 0.938 \\
    Qwen3-VL-4B-Instruct            & 0.965 & 0.968 & 0.963 \\
    Qwen3-VL-4B-Thinking            & 0.965 & 0.968 & 0.960 \\
    Qwen3-VL-8B-Instruct            & 0.968 & 0.970 & 0.963 \\
    Qwen3-VL-8B-Thinking            & 0.969 & 0.972 & 0.964 \\
    \bottomrule
  \end{tabular}
\end{table}

\subsection{Stability of the Model--Human Gap}
\label{app:embedding_gap}

Beyond the consistency of the scores themselves, we examined whether our main conclusions hold with different choices of embedding models. Using Qwen3-Embedding-0.6B and gemini-embedding-001, we repeated the Top-1 Accuracy analysis (Fig.~\ref{fig:overall_eval}a) and the four attribute sensitivity analyses (Fig.~\ref{fig:bias}), with human--human consistency and the human sensitivity baselines recomputed in the corresponding embedding space.

\begin{figure}[ht]
  \centering
  \begin{subfigure}[b]{0.48\linewidth}
    \centering
    \includegraphics[width=\linewidth]{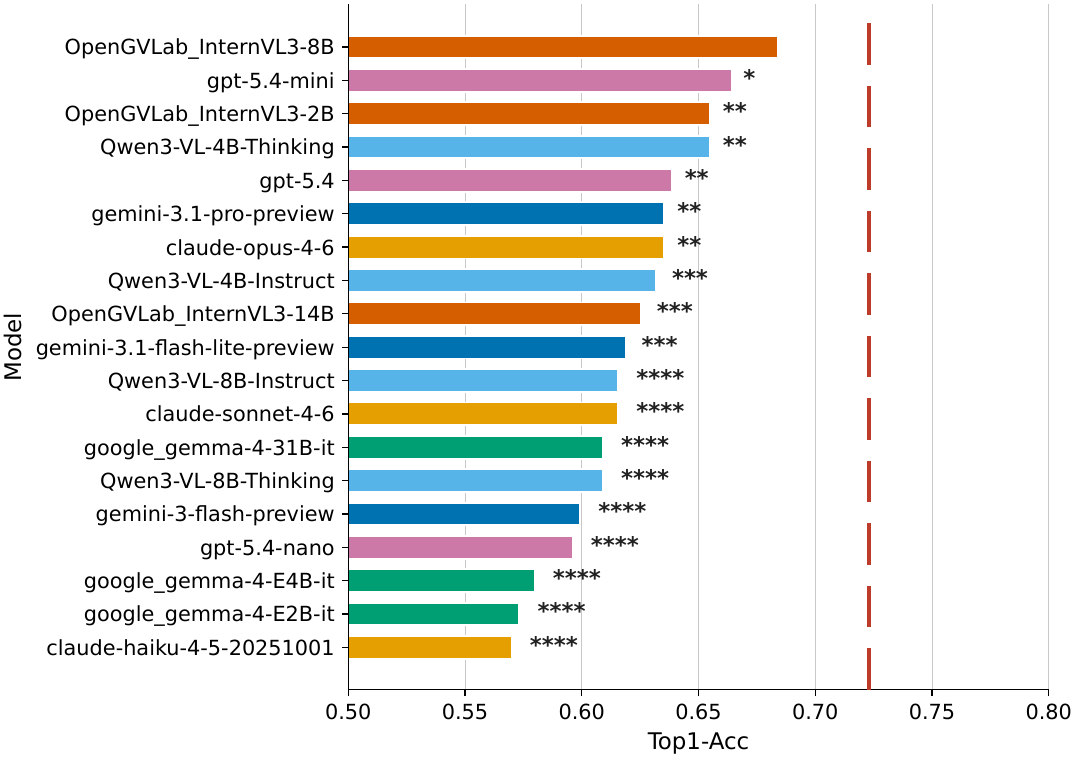}
    \caption{Qwen3-Embedding-0.6B}
    \label{fig:embedding_acc_qwen}
  \end{subfigure}
  \hfill
  \begin{subfigure}[b]{0.48\linewidth}
    \centering
    \includegraphics[width=\linewidth]{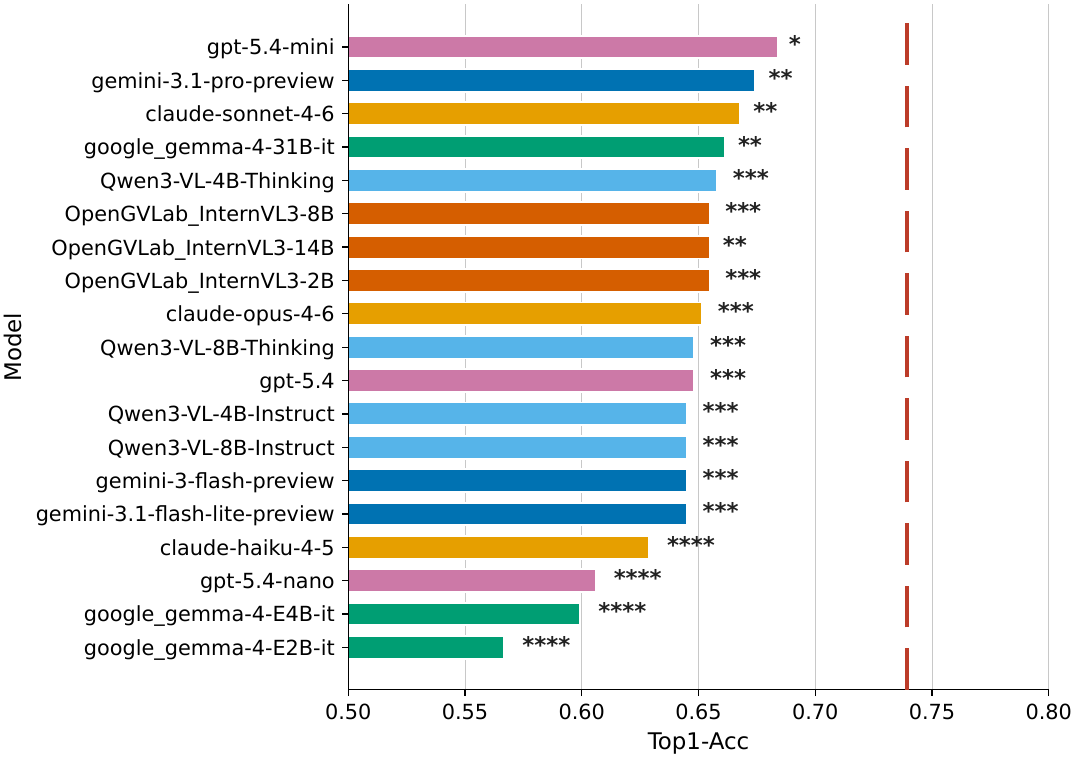}
    \caption{gemini-embedding-001}
    \label{fig:embedding_acc_gemini}
  \end{subfigure}
  \caption{Top-1 Accuracy at predicting the critical object when CSS is computed with different sentence embedding models. The dashed line denotes human--human consistency computed with the same embedding model.}
  \label{fig:embedding_acc}
\end{figure}

\paragraph{Accuracy gap.}
As shown in Fig.~\ref{fig:embedding_acc}, the systematic gap between human--human consistency and model Top-1 Accuracy persists under both alternative embedding models, consistent with the results obtained using Jasper.

\begin{figure}[ht]
  \centering
  \begin{subfigure}[b]{0.48\linewidth}
    \centering
    \includegraphics[width=\linewidth]{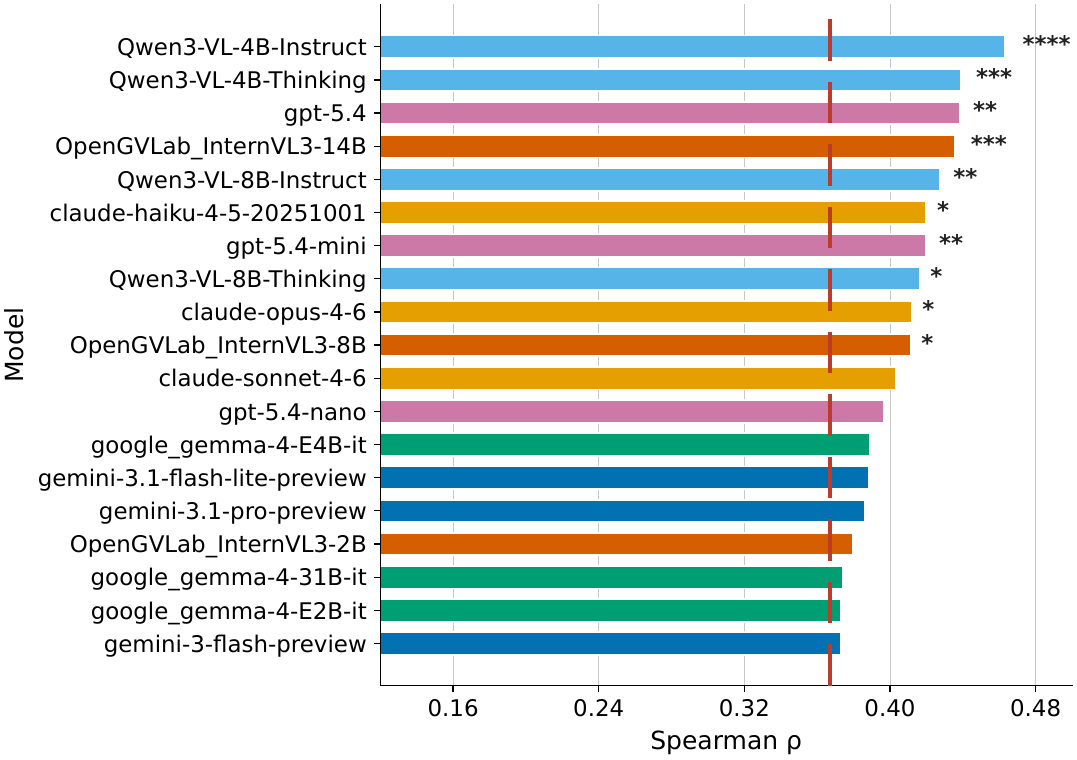}
    \caption{Sensitivity to Size}
  \end{subfigure}
  \hfill
  \begin{subfigure}[b]{0.48\linewidth}
    \centering
    \includegraphics[width=\linewidth]{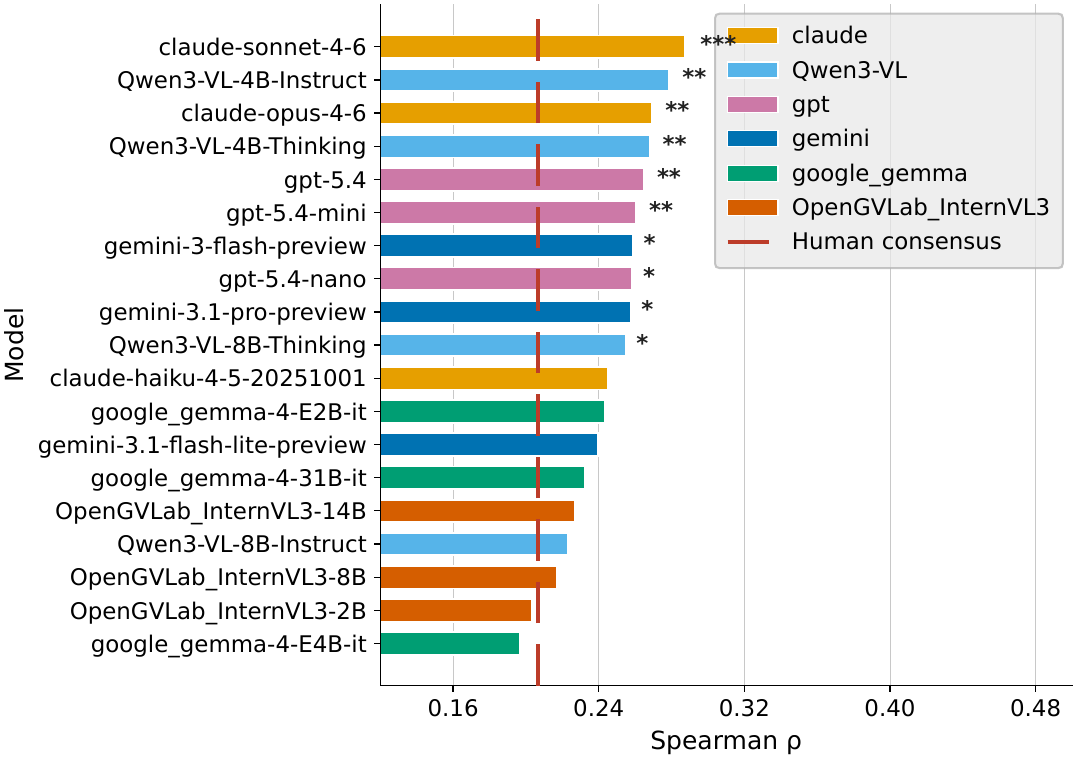}
    \caption{Sensitivity to Center}
  \end{subfigure}

  \vspace{0.5em}

  \begin{subfigure}[b]{0.48\linewidth}
    \centering
    \includegraphics[width=\linewidth]{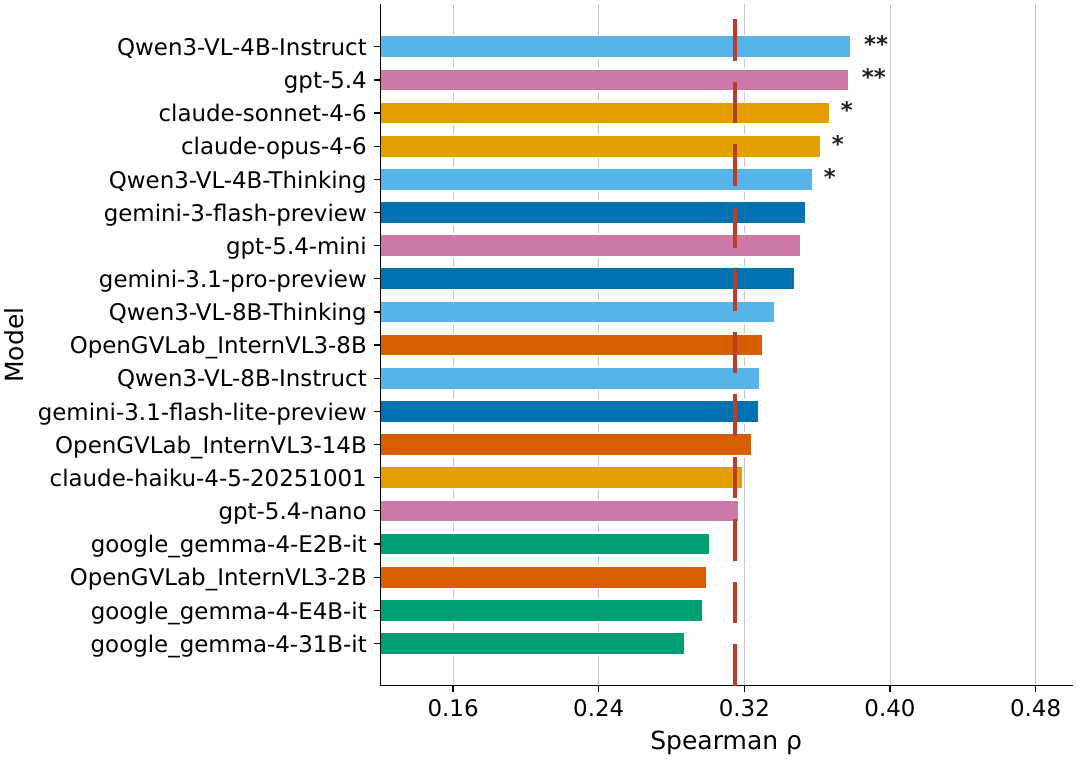}
    \caption{Sensitivity to Low-level Saliency}
  \end{subfigure}
  \hfill
  \begin{subfigure}[b]{0.48\linewidth}
    \centering
    \includegraphics[width=\linewidth]{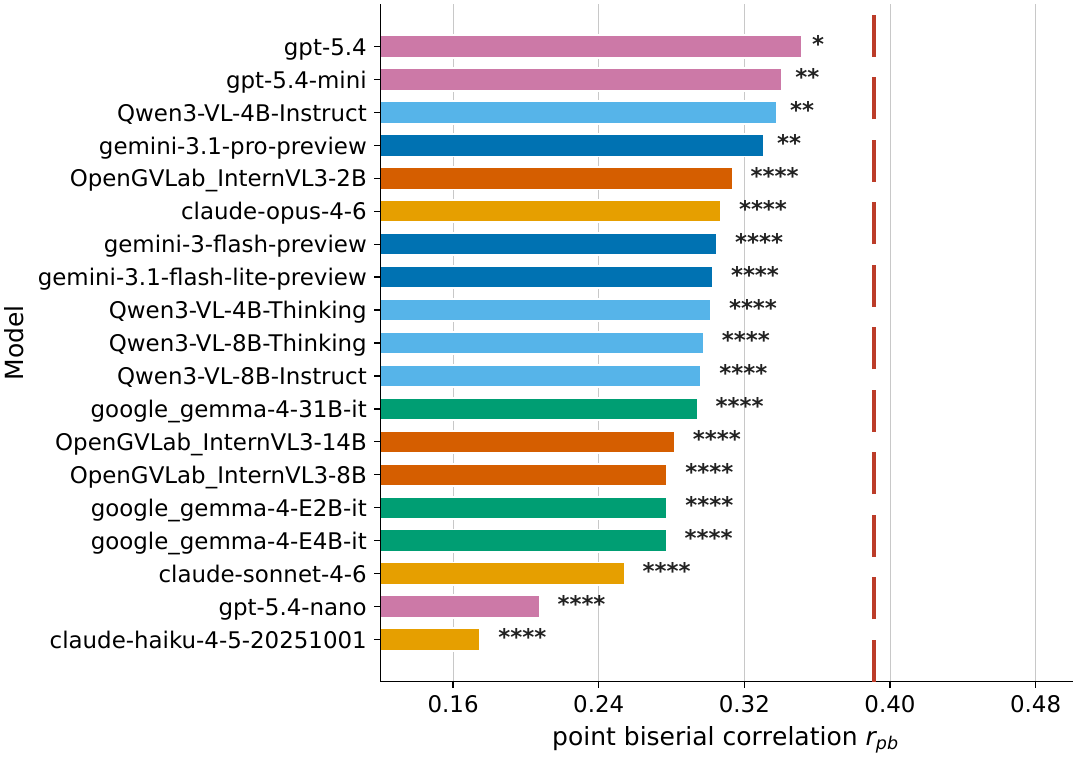}
    \caption{Sensitivity to Person-Non-person}
  \end{subfigure}
  \caption{Attribute Sensitivity of VLMs when CSS is computed with Qwen3-Embedding-0.6B.}
  \label{fig:embedding_bias_qwen}
\end{figure}

\begin{figure}[ht]
  \centering
  \begin{subfigure}[b]{0.48\linewidth}
    \centering
    \includegraphics[width=\linewidth]{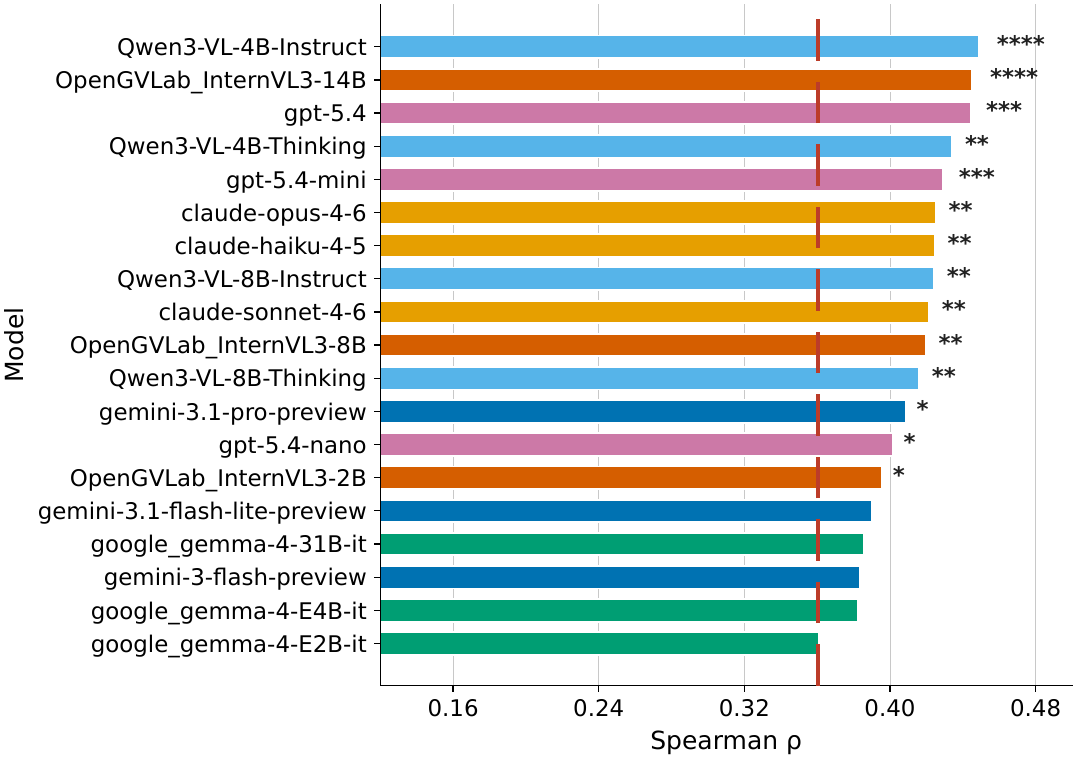}
    \caption{Sensitivity to Size}
  \end{subfigure}
  \hfill
  \begin{subfigure}[b]{0.48\linewidth}
    \centering
    \includegraphics[width=\linewidth]{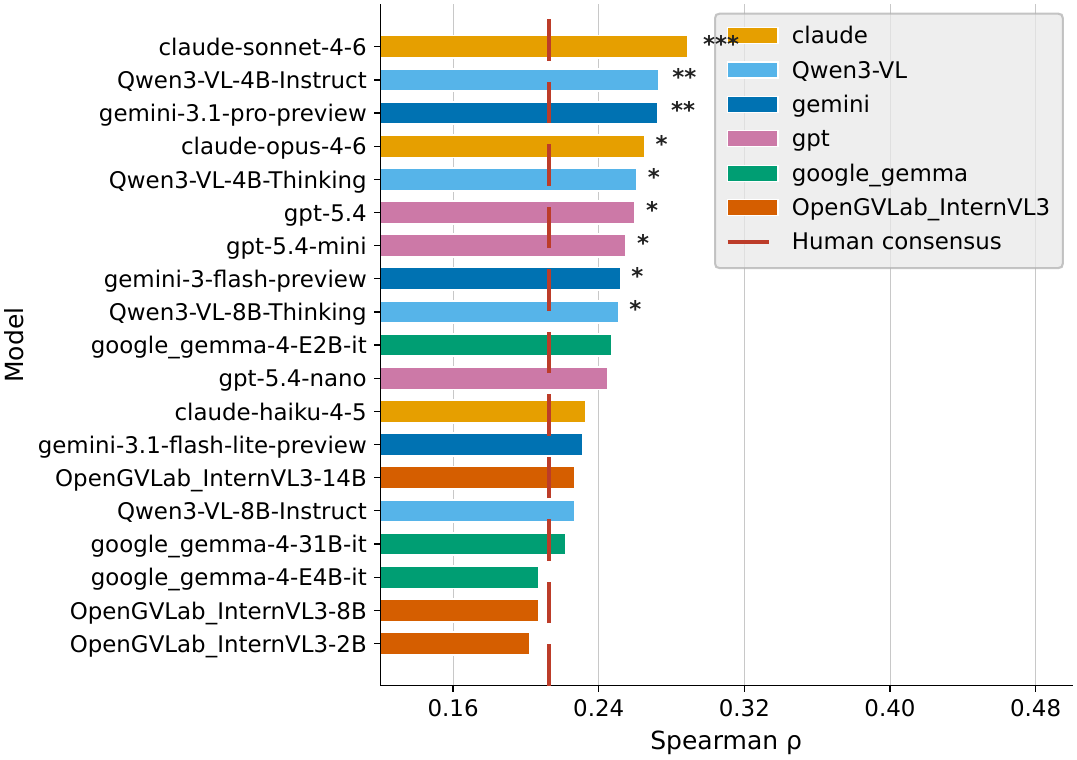}
    \caption{Sensitivity to Center}
  \end{subfigure}

  \vspace{0.5em}

  \begin{subfigure}[b]{0.48\linewidth}
    \centering
    \includegraphics[width=\linewidth]{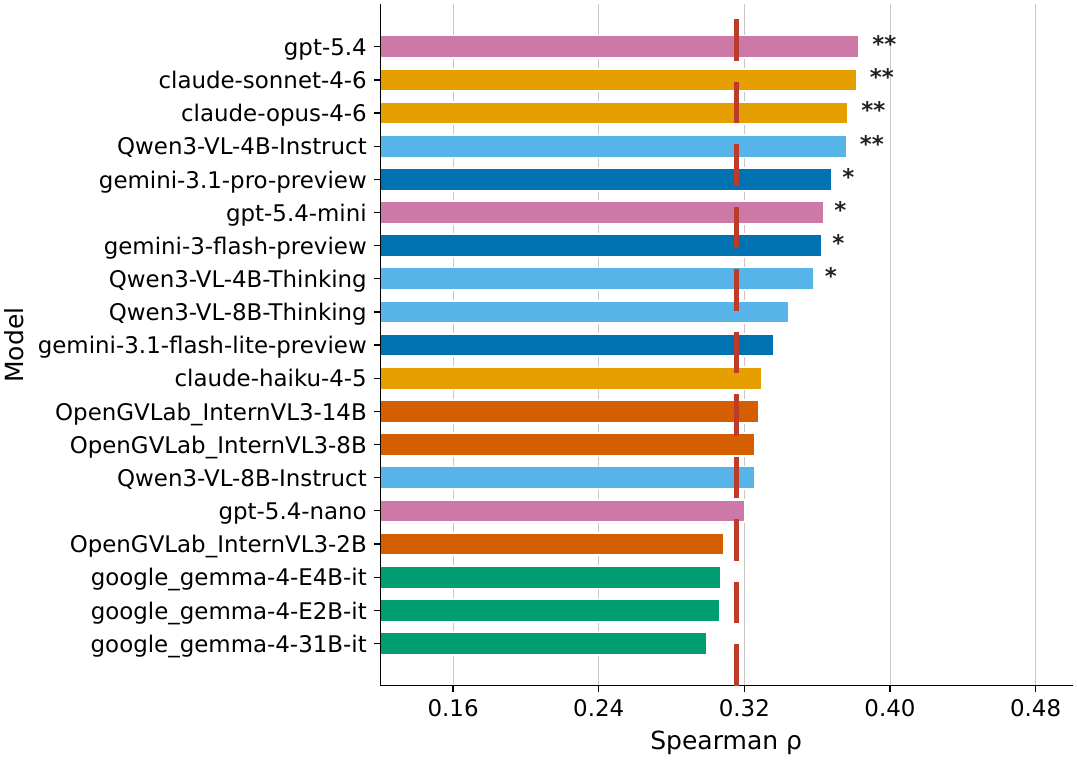}
    \caption{Sensitivity to Low-level Saliency}
  \end{subfigure}
  \hfill
  \begin{subfigure}[b]{0.48\linewidth}
    \centering
    \includegraphics[width=\linewidth]{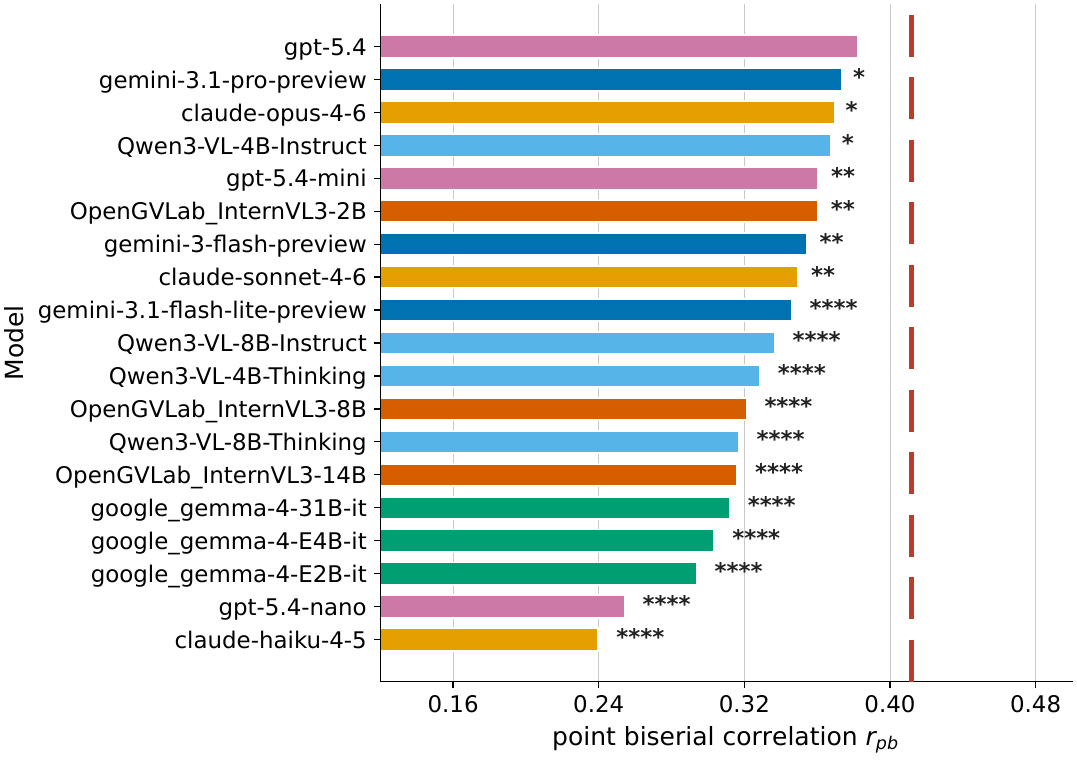}
    \caption{Sensitivity to Person-non-person}
  \end{subfigure}
  \caption{Attribute Sensitivity of VLMs when CSS is computed with gemini-embedding-001.}
  \label{fig:embedding_bias_gemini}
\end{figure}

\paragraph{Attribute Sensitivity gap.}
Fig.~\ref{fig:embedding_bias_qwen} and Fig.~\ref{fig:embedding_bias_gemini} report the four attribute sensitivities under Qwen3-Embedding-0.6B and gemini-embedding-001, respectively.
Together with the high consistency of CSS scores across embedding models, these results show that our findings are not artifacts of a particular sentence embedding model. In both cases, the divergences observed in the main text are mostly preserved: compared with humans, models are more sensitive to Size, Center, and Low-level Saliency, and insensitive to whether the object is a person or not, except for the sensitivity to low-level saliency measured under Qwen3-Embedding-0.6B, where there remain fewer models that are significantly different from the human baseline. 

\section{Variability in Human Agreement Does Not Account for the Model--Human Divergence}
\label{app:human_certainty}

Not every scene necessarily has a robust human consensus. For some images, observers may disagree about which object is most critical to the scene, so the ground truth itself is uncertain. One possibility is that the gap between humans and models is due to the variability among human observers rather than a divergence in scene comprehension. Here, we test whether the gap persists after controlling for the certainty of human agreement.

\paragraph{Quantifying human certainty.}
Because our psychophysics experiment follows a between-subjects design, each counterfactual variant of a scene was described by a different group of participants. Individual differences therefore cannot be measured directly. Instead, we estimate the certainty of human agreement for each scene as follows. For each condition of a scene (the factual image and each of its counterfactual variants), we randomly select a single participant's description from the 10 participants assigned to that condition. Using these single-participant descriptions, we compute the CSS of each object and identify the critical object as the one with the highest CSS. We repeat this sampling procedure and define the human certainty score of the scene as the probability that the critical object obtained from single participants matches the critical object determined from all participants' responses. A higher score indicates that the identification of the critical object is consistent across individual observers, whereas a lower score indicates that the human consensus is less stable.

\paragraph{Results.}
We split the 307 factual scenes into four quartile bins according to their human certainty scores (77 scenes per bin, except Q3, which contains 76). Within each bin, we report the Top-1 Accuracy averaged across all 19 evaluated VLMs alongside the human--human consistency, computed using the same procedure described in Sec.~\ref{sec:Perceptual_Gap}. Error bars denote the standard deviation of the bootstrap distribution, and statistical significance is assessed with a one-tailed bootstrap test of whether human--human consistency exceeds the mean model accuracy.

\begin{table}[ht]
  \caption{Model--human gap in critical object identification across levels of human certainty. Scenes are split into quartiles based on the certainty of human agreement. Mean Model Top-1 Acc is averaged across all 19 evaluated VLMs. Values are reported as mean $\pm$ bootstrap standard deviation; $p$-values are obtained from one-tailed bootstrap tests, indicating whether the human--model gap is significant.}
  \label{tab:human_certainty}
  \centering
  \small
  \begin{tabular}{lcccc}
    \toprule
    Certainty Bin & Certainty Range & Mean Model Top-1 Acc & Human--Human Consistency & Gap p \\
    \midrule
    Q1 & $[0.226, 0.462]$ & $0.338 \pm 0.029$ & $0.286 \pm 0.052$ & $0.807$ \\
    Q2 & $(0.462, 0.635]$ & $0.506 \pm 0.037$ & $0.649 \pm 0.055$ & $0.008$ \\
    Q3 & $(0.635, 0.850]$ & $0.735 \pm 0.026$ & $0.987 \pm 0.013$ & $<0.001$ \\
    Q4 & $(0.850, 1.000]$ & $0.918 \pm 0.018$ & $1.000 \pm 0.000$ & $<0.001$ \\
    \bottomrule
  \end{tabular}
\end{table}

As shown in Table~\ref{tab:human_certainty}, models and humans show no significant difference only in the lowest-certainty bin (Q1). However, in all three remaining bins, human--human consistency significantly exceeds the mean model accuracy. Notably, the gap remains significant even in Q2, and it persists through the highest-certainty bins (Q3 and Q4), indicating that the model--human divergence cannot be explained simply by the variability of human agreement.


\section{Extended Results: Model-Human Ranking Correlations}

\begin{figure}[hp]
    \centering
    \includegraphics[width=0.85\linewidth]{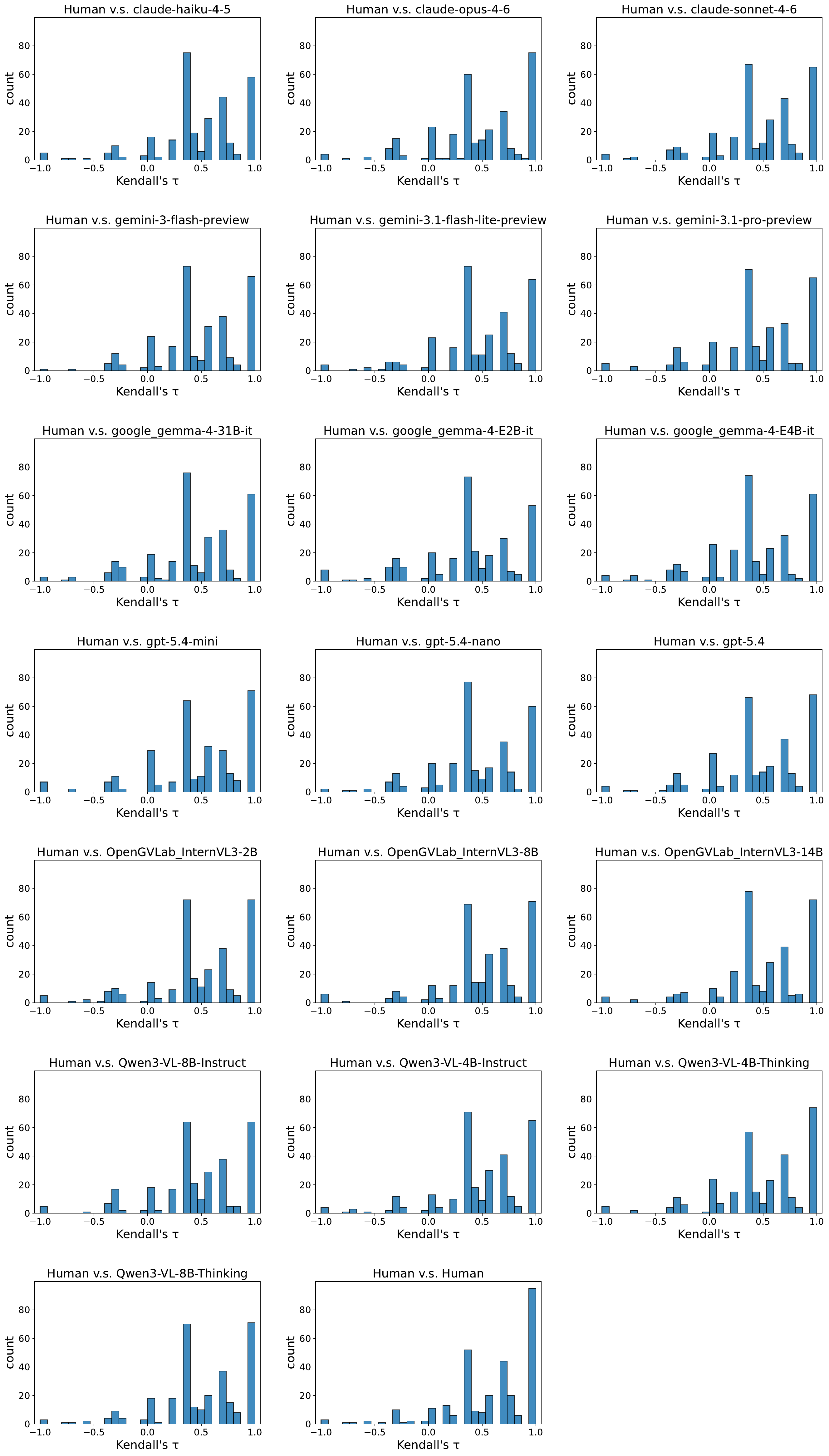}
    \caption{Distribution of Kendall's $\tau$  for each model comparing against human, and human-human consistency}
    \label{fig:grid_kendall}
\end{figure}

Due to space constraints, the main text reports only the aggregated mean Kendall's $\tau$ coefficients for the object importance rankings between models and human observers across the dataset. However, because Kendall's $\tau$ is computed independently for each factual scene, examining the full distribution of these coefficients provides a much more granular perspective on model--human alignment. In this section, we present the complete distributions of Kendall's $\tau$ for each evaluated Vision-Language Model (VLM) when compared against the human baseline (Fig.~\ref{fig:grid_kendall}). The human--human consistency is also illustrated at the end of the figure (computed in the same way as described in Sec.~\ref{sec:Perceptual_Gap}).

\section{Extended Qualitative Comparisons: CSS vs. White-Box Saliency}

\begin{figure}[hp]
    \centering
    \includegraphics[width=\linewidth]{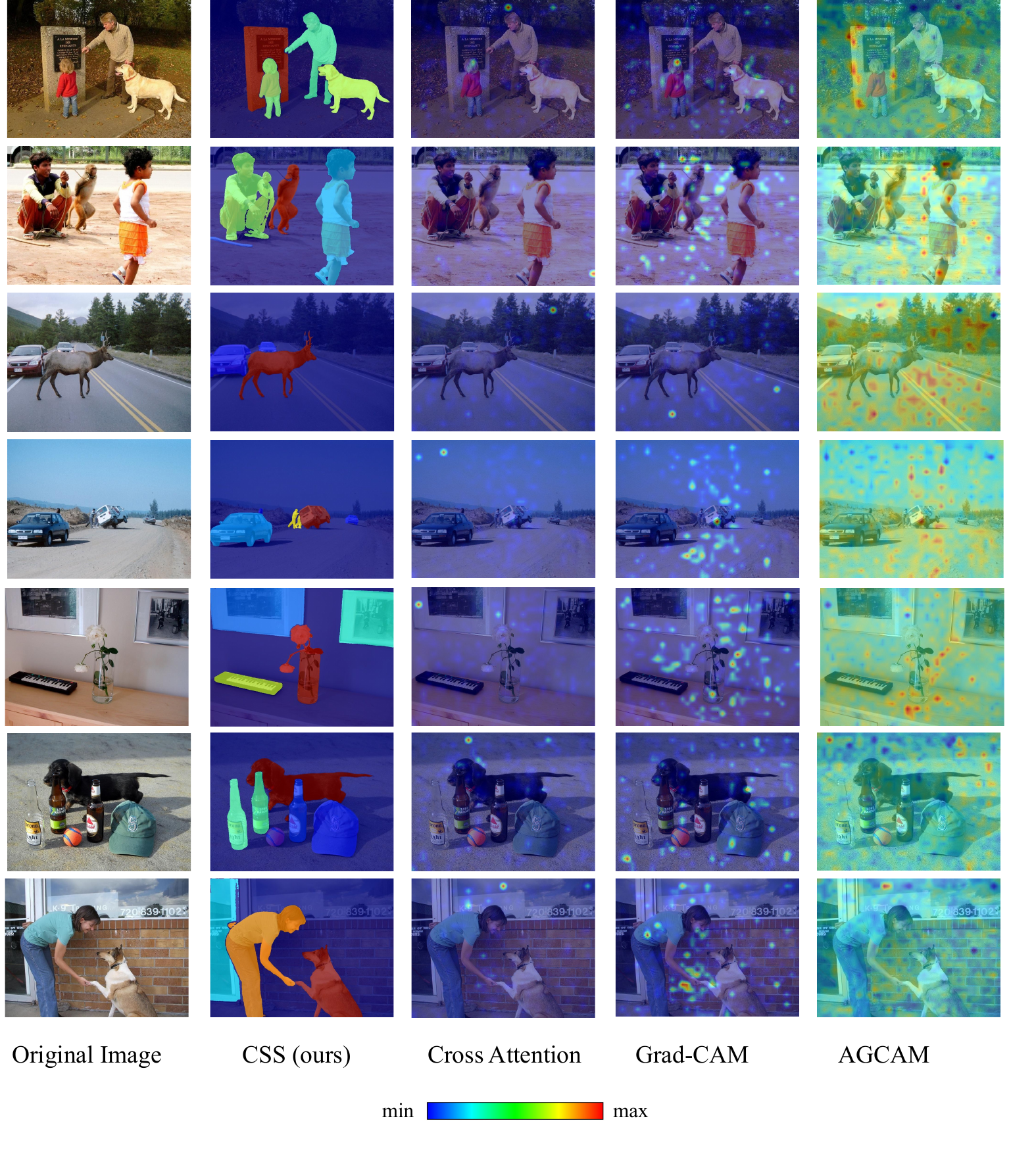}
    \caption{Comparisons between CSS Maps and attention-based white-box maps}
    \label{fig:whitebox_extended}
\end{figure}

In this section, we provide extended qualitative visualizations comparing white-box interpretability methods against our causal Counterfactual Semantic Saliency (CSS) framework (Fig.\ref{fig:whitebox_extended}). The activation values within the CSS maps are globally normalized across all factual scenes in our dataset to standardize the scale of the induced semantic shift.

\section{Broader Impacts}

Understanding the divergence in scene perception is critical for AI safety. We reveal that modern VLMs overly rely on spatial dominance rather than top-down semantic importance (such as the presence of humans). Identifying and mitigating this unnatural sensitivity is crucial before deploying VLMs in high-stakes environments, such as autonomous driving or medical imaging, where ignoring a small but semantically critical object (e.g., a pedestrian) in favor of a large background element could have catastrophic consequences. Conversely, exposing these specific mechanistic vulnerabilities could be exploited by malicious actors. By demonstrating the severity of VLM's reliance to Size and Center, adversarial attacks could be designed using physically large, irrelevant distractors to intentionally blind AI security systems to actual targets.



\end{document}